\documentclass[11pt]{article}
\usepackage[margin=1in]{geometry}
\usepackage{amsmath,amssymb,amsthm}
\usepackage{booktabs}
\usepackage{graphicx}
\graphicspath{{figures/}{./figures/}}
\usepackage{placeins}
\usepackage{afterpage}
\usepackage{hyperref}
\usepackage{xcolor}
\usepackage{enumitem}
\usepackage{algorithm}
\usepackage{algpseudocode}
\usepackage{microtype}
\usepackage{xurl}
\hypersetup{colorlinks=true, linkcolor=blue!50!black, citecolor=blue!50!black, urlcolor=blue!50!black}

\title{\textbf{AI Trading's Alpha Singularity:
Emergent Market Reasoning through Agent-to-Agent Self-Evolution}}

\author{\textbf{Yuqi Li} \\
  \textbf{Panda AI} \\
  \texttt{libubai@pandaai.online} \\\and
  \textbf{Siyuan Liu} \\
  \textbf{Panda AI} \\
  \texttt{liusiyuan@pandaai.online} \\\and
  \textbf{Bingjun Liu} \\
  \textbf{Panda AI} \\
  \texttt{liubingjun@pandaai.online} \\}
\date{}

\begin{document}
\maketitle

\begin{abstract}
Automated alpha mining holds the scoring function fixed and varies the
search algorithm over it. A search that converges against a fixed
scorer overfits whatever the scorer cannot penalize, which we argue is
a primary cause of the out-of-sample generalization gap.
We treat the scoring function as a search artifact alongside the alpha
factors and study what conditions make this joint search admissible.
\textbf{Sealed Joint Search (SJS)} is the framework we propose: a
small set of structural conditions on the information flow inside an
autonomous-discovery system that prevent the joint search from
collapsing into self-confirmation while keeping the external evaluator
sealed. The conditions cover role decomposition, typed inter-role
communication, provenance-sealed reads, persistent versioned artifact
stores, and a substrate-local promotion rule.
\textbf{Agora} is the system we build to test SJS empirically: five
role-specialized LLM agent classes communicating through three typed
channels, jointly evolving eight skill libraries, with the alpha-side
libraries built on the AlphaGen operator vocabulary. Three independent
evaluator instances write narrative reports that the orchestrator
aggregates into a single brief, so adjudication carries forward
disagreement rather than collapsing it to a vote.
We run Agora for 100 outer rounds on CSI~1000 and evaluate on a 91-day
2026 holdout sealed from every LLM-facing input. Agora reaches a
holdout portfolio Sharpe of $+1.87$; the strongest baseline reaches
$+1.334$ at a favorable seed and $-0.755$ on the cross-seed mean.
Pre-loading the two metrics that Agora discovers into a frozen-library
ablation recovers only $+0.40$ of the $+2.25$ Sharpe gap, and adding
PPO without library evolution makes the gap worse. The two metrics
behave like emergent properties of the system rather than designed
components.
Two caveats: the full Agora run is single-seed, and the signal is
short-side concentrated and therefore intended for long-short
deployment.
\end{abstract}

\section{Introduction}
\label{sec:introduction}

Autonomous discovery problems share a common shape: a search procedure
proposes candidate artifacts, a scoring function ranks them, and an
external evaluator adjudicates the result on data the search never
touched. Quantitative alpha mining is one instance. Autonomous theorem
proving, autonomous program synthesis, and autonomous experimental
design are others. In alpha mining, the methods used to find
return-predictive expressions have evolved through three generations:
hand-crafted factor models (Fama--French
\cite{fama1993,fama2015}, momentum \cite{jegadeesh1993,carhart1997},
the Alpha101 catalog \cite{kakushadze2016}); symbolic regression and
genetic programming
\cite{koza1992,schmidt2009,lin2019autoalpha,cui2021alphaevolve}; and
RL-based search that casts the expression tree as a sequential
decision \cite{yu2023alphagen,schulman2017ppo,petersen2021dsr}.

Across these three generations and across other autonomous-discovery
domains, the scoring function is fixed at design time. The search
procedure cannot detect when its own objective has been overfit to
the training segment, and any reported gain is conditional on a
hyperparameter that no participant revises. With the multiple-testing concerns documented in
\cite{bailey2014, harvey2016, kaufman2012leakage,lopezdeprado2018},
this is a sharp constraint: when the bottleneck is the choice of
objective rather than the choice of expression, better search
procedures cannot help.

Making the scoring function itself a search artifact is one response. Existing self-evolving systems
\cite{wang2023voyager,ma2023eureka,romera2024funsearch,hu2024adas,lu2024aiscientist,zelikman2022star}
address the resulting collapse in two ways: either by holding a
criterion fixed (and inheriting its blind spots) or by querying a
strong external evaluator on every proposal (and paying the
sample-cost and leakage cost that follow). Both routes leave the
joint-search problem itself unaddressed.

This paper introduces a third regime: \textbf{Sealed Joint Search
(SJS)}. The criterion is searched, the external evaluator is sealed,
and the connection between them is mediated by substrate-level
constraints that no participating agent controls. SJS is a set of
structural conditions on the search procedure, not a system. The
five conditions are: F1 decomposed proposal, in which proposers and
adjudicators are separate roles with no shared state; F2 typed
inter-role communication, with no free-form chat between roles; F3 a
provenance-sealed substrate, in which every record is tagged with
its data segment and reads are filtered by role and content type; F4
persistent versioned artifact stores, role-owned, with a state
machine over builtin / trial / accepted / rejected; and F5 a
substrate-local promotion rule that is non-self-judging and
outcome-grounded. A boundary condition fixes the external evaluator
before the run and keeps it inaccessible to every role.

To validate SJS empirically, we instantiate it as \textbf{Agora}, an
A2A LLM system in which five role-specialized agent classes (nine
clients with isolated contexts) communicate through three typed
channels and jointly evolve eight skill libraries on a sealed
substrate. The substrate routes each agent's output into a shared typed record that later agents read without seeing each other's contexts. Adjudication in this realization is
deliberation: independent evaluator instances produce narrative
reports that the substrate aggregates rather than reducing to a
single vote, so dissenting opinion enters the next round as evidence.
The alpha-side libraries (operator, RL-network,
RL-algorithm, reward) are built on AlphaGen \cite{yu2023alphagen};
the metric, topic, rubric, and meta-rubric libraries are seeded with
deliberately sparse builtin entries. Promotion runs against
training-segment Sharpe with $|\rho| > 0.3$. The 2026 holdout is
loaded once after the run completes and is sealed from every
LLM-facing input.

The central research question is the decomposition of holdout
performance under an SJS realization: how much of any out-of-sample
gain is attributable to (a) library evolution, (b) the A2A
decomposition itself, (c) the LLMs in isolation, and (d) the
underlying RL or symbolic search machinery. We address this through
a 100-round Agora run on CSI~1000 against seven baselines (genetic
programming, pure PPO, single-LLM one-shot, iterative single-LLM,
Alpha101, a frozen-library ablation, and random search) plus three
within-Agora ablations.

\paragraph{Contributions.}
\begin{itemize}
\item \textbf{Sealed Joint Search (SJS),} a methodological framework
  for joint search over signals and scoring functions under a sealed
  external evaluator. The framework consists of five structural
  conditions (F1--F5) on the information flow of the search procedure
  plus a boundary condition on the evaluator. SJS is independent of
  any particular optimizer, language model, or operator language.
\item \textbf{Agora,} an instantiation of SJS as an A2A LLM system,
  satisfying F1--F5 with five role-specialized agent classes, three
  typed channels, eight skill libraries, and an empirical promotion
  rule against train-segment Sharpe. Agora is evaluated on a sealed
  91-day 2026 holdout of CSI~1000.
\item \textbf{Empirical evidence about SJS, demonstrated on Agora.}
  Agora reaches holdout portfolio Sharpe $+1.87$, ahead of seven
  baselines on point estimates. Across 100 rounds, evolution emerged
  in exactly one of the eight skill libraries (metric), with two
  promoted entries that the system was not designed to produce.
  Neither metric was produced by any single agent: each emerged from
  aggregate promotion evidence across rounds.
  Ablation isolates an upper bound of
  $+2.25$ Sharpe units of total contribution from the joint
  F4 + F5 mechanism (persistence + substrate-local promotion), of
  which the static value of two LLM-discovered metrics accounts for
  $+0.40$. PPO relay added to a frozen system is harmful, indicating
  that asymmetric upgrades of the search procedure without
  co-evolving scoring are counterproductive.
\item \textbf{Open-source release.} We release Agora's implementation,
  all seven baseline configurations under a common evaluation
  harness, the leakage audit, and the per-round registry snapshots
  for the eight skill libraries.
\end{itemize}

\section{Related Work}
\label{sec:related_work}

\paragraph{Symbolic alpha generation.}
Three generations of methods have searched the space of arithmetic
expressions over OHLCV variables for ones that correlate with
forward returns. Hand-crafted formulas
\cite{kakushadze2016,fama1993,fama2015,carhart1997,jegadeesh1993}
predate any automated search; AutoAlpha \cite{lin2019autoalpha} and
AlphaEvolve \cite{cui2021alphaevolve} introduced evolutionary search
over operator trees \cite{koza1992,schmidt2009}, and AlphaGen
\cite{yu2023alphagen} replaced the genetic engine with a Maskable
PPO policy \cite{schulman2017ppo,raffin2021sb3}. Deep symbolic
regression \cite{petersen2021dsr} forms a parallel line within
symbolic search. Across these methods, the fitness function
(typically rank IC against a forward return) is fixed at design
time and only the search procedure over expression trees varies.
Our baselines B1 (a gplearn adaptation of the AlphaGen reference)
and B2 (AlphaGen pure PPO) are drawn from the most recent two
generations. The multiple-testing concerns documented in
\cite{harvey2016,bailey2014} motivate the conservative attribution
strategy we adopt in Section~\ref{sec:results}.

\paragraph{LLMs for code generation in finance.}
Recent work prompts language models to emit alpha
expressions or trading code, building on the general code-generation
literature
\cite{chen2021codex,vaswani2017,brown2020gpt3,openai2023gpt4,anthropic2024claude3}
and on finance-tuned models
\cite{wu2023bloomberggpt,yang2023fingpt}. FinMem
\cite{li2023finmem} layers structured memory on top of a single
agent, and TradingAgents \cite{xiao2024tradingagents} arranges
multiple LLM roles around a trading task. These systems hold the
scoring function fixed and use the LLM to write better $f$. We
include the simplest two variants of this setup, single-LLM
one-shot (B3) and single-LLM iterative against IC feedback (B4).
Agora differs from all of these: it treats the scoring function itself as a search artifact
under an empirical promotion rule, not as a fixed objective the LLM is asked to optimize.

\paragraph{Self-improving LLM agents.}
Outside finance, several systems let an LLM edit its own toolset.
Voyager builds a Minecraft skill library
\cite{wang2023voyager}, Eureka writes RL reward functions
\cite{ma2023eureka}, ADAS searches over agent designs
\cite{hu2024adas}, FunSearch discovers mathematical constructions
\cite{romera2024funsearch}, the AI Scientist runs an end-to-end
research loop \cite{lu2024aiscientist}, and STaR
\cite{zelikman2022star} bootstraps reasoning chains from a model's
own outputs. The reasoning and self-correction building blocks
\cite{yao2023react,yao2023tot,wei2022cot,shinn2023reflexion,madaan2023selfrefine}
and tool-use mechanisms
\cite{schick2023toolformer,wang2024codeact} are studied separately.
Agora adopts the basic shape (LLM proposes Python, system sandboxes
and validates, accepted code joins the runtime), and adds two
elements specific to alpha discovery: a sealed evaluation
substrate that distinguishes the segment used to evolve metrics
from the segment used to evaluate them, and an empirical promotion
rule that admits a candidate metric only when its predictions
correlate with realized train Sharpe.

\paragraph{Multi-agent and agent-to-agent LLM systems.}
Several frameworks coordinate multiple LLM instances on a shared
task: MetaGPT models a software-development hierarchy
\cite{hong2024metagpt}, AutoGen provides a general conversation-based
framework \cite{wu2024autogen}, CAMEL \cite{li2023camel} and ChatDev
\cite{qian2024chatdev} use role-playing dialogues, and Generative
Agents \cite{park2023generative} demonstrate emergent dynamics over
a shared memory. AgentBench \cite{liu2023agentbench} evaluates
role-specialized agents on a range of tasks, and capability surveys
\cite{liang2023cap} catalog the design space. The coordination
substrate in most of these systems is a chat history that the
participating agents read directly. We use the term agent-to-agent
(A2A) for the narrower setting in which each agent runs on an
isolated context (its own client, message history, and system
prompt) and all cross-agent information is forced through a typed
substrate that the agents do not control. Agora is A2A in this
sense: the LLM Wiki and the directed advisory bundles are typed
channels with a role-scoped read filter, and no agent can read
another's full context. With this isolation, the substrate filters
holdout-segment information at the channel level rather than by
instruction; shared-history multi-agent designs cannot enforce
that constraint.

\paragraph{Statistical inference and out-of-sample evaluation.}
The data-leakage and out-of-sample-evaluation literature predates
the LLM-agent literature and supplies the inferential tools we use:
Newey--West HAC standard errors \cite{newey1987} for serially
correlated daily returns, the reality-check and data-snooping
framework of \cite{white2000}, the predictive-accuracy test of
\cite{diebold1995}, and the leakage taxonomy of
\cite{kaufman2012leakage,arnold2024datacontamination}. The deflated
Sharpe and backtest-overfitting diagnostics of
\cite{bailey2014,lopezdeprado2018} inform our reporting choices.
In the LLM-agent setting, contamination is a structural property
of any pipeline in which agents can read each other's contexts;
forbidding agents from reading the holdout by instruction is not
sufficient, and the substrate must enforce the constraint.

\paragraph{Open-ended search.}
Expanding a system's search vocabulary can outperform optimizing
within a fixed one; this observation has a long history in
evolutionary computation
\cite{lehman2008novelty,stanley2017open,wang2019paired}. Agora
inherits the basic idea, in that artifacts the system produces
(new metrics, new operators, new rubrics) become part of the search
space for later rounds, but applies it under the sealing constraints
imposed by financial out-of-sample evaluation
\cite{lopezdeprado2018}. Open-ended vocabulary expansion under a sealed evaluation segment does not appear in this literature.

\section{Methodology}
\label{sec:method}

\subsection{Methodological Reorientation}
\label{subsec:lens}

For thirty years, the methodology of automated discovery has had a
fixed shape. A search algorithm proposes candidate artifacts, a
scoring function rates them, and a held-out evaluator decides whether
the search produced something of value. Methodological progress
means improving the search algorithm. The scoring function is
inherited from prior work, and the held-out evaluator is a passive
recipient of the final output. Genetic programming, reinforcement
learning, and prompt-engineered language-model proposers are all
methodologically commensurable in this sense: they are different
search algorithms operating on the same three-component template.

We argue that this template has run out of methodological room.
Improvements to the search algorithm under a fixed scoring function
amount to faster exploitation of the scoring function's blind spots.
The empirical evidence is uniform across decades of factor-mining,
program-search, and policy-learning literature: when the scoring
function is held fixed, the search converges to artifacts that are
maximally aligned with whatever the scoring function fails to
penalize \cite{bailey2014, harvey2016, kaufman2012leakage}. The
limit of search-algorithm research, under a fixed scorer, is the
limit of overfitting that scorer.

Making the scorer itself a learned object appears in pieces in the
recent self-evolving-LLM literature \cite{wang2023voyager,
ma2023eureka, romera2024funsearch, hu2024adas, lu2024aiscientist,
zelikman2022star} and in the multi-agent-LLM literature
\cite{hong2024metagpt, wu2024autogen, li2023camel, qian2024chatdev,
park2023generative}. The pieces have not been combined into a
methodology, because the obstacle is not mechanical. It is
conceptual. A search procedure that proposes both the artifact and
the scorer collapses trivially: any procedure with control over both
objects can satisfy itself by relaxing the second to admit the
first. Multiplying agents, layering critique loops, or querying a
stronger evaluator at training time do not address the collapse;
they relocate it.

This obstacle dissolves under a methodological
reorientation. The object of methodological study is no longer the
search algorithm. It is the \emph{information topology} of the
search: which roles produce information of which type, which roles
are permitted to consume it, what fidelity it has when it crosses a
boundary, and what state persists across the boundary. Under this
reorientation, generalization is a property of the topology, not of
any participant's behavior. The substrate becomes a first-class
methodological concern, on equal footing with the optimizer that
runs on it.

The methodology we propose, \emph{Sealed Joint Search} (SJS), is the
analytical framework that follows from this reorientation. SJS does
not specify a search algorithm. It specifies the information
topology under which a joint search over $(f, g)$ pairs admits a
non-degenerate solution. Section~\ref{subsec:topology} formalizes
the framework and Section~\ref{subsec:sealed-topology} states the
five topological properties that characterize the sealed regime.
Section~\ref{subsec:a2a} introduces the
\emph{agent-to-agent} (A2A) realization of SJS, which instantiates the framework
on existing language models. Section~\ref{subsec:agora} introduces
Agora, the A2A system on which we report empirical evidence;
further implementation detail is in Section~\ref{sec:experiments}.

\subsection{Information-Topology Formulation of Joint Search}
\label{subsec:topology}

Let $\mathcal{F}$ be a space of candidate artifacts and $\mathcal{G}$
a space of candidate scoring functions. Let $u$ be an external
evaluator on a sealed segment $\mathcal{D}_{\text{holdout}}$. Joint
search optimizes
\[
  (f^\star, g^\star) = \arg\max_{(f,g)} \; u(f, \mathcal{D}_{\text{holdout}}),
\]
under structural constraints on which we focus.

A search procedure is, abstractly, a directed graph. Its nodes are
\emph{loci}: producers (of $f$, of $g$), adjudicators, and stores of
artifacts. Its edges are typed information channels, each labeled
with a content type, a fidelity, and a provenance tag. Its temporal
structure is a schedule that determines when each node may write to
each outgoing edge and when each may read from each incoming edge.
Under this view, the search algorithm is the local update rule at a
single node; the scoring function is the local update rule at another.
Generalization is determined by the global graph, not by any single rule.

The collapse described in Section~\ref{subsec:lens} is a topological
property: it occurs whenever the producer of $f$ and the producer of
$g$ share an unfiltered edge. Under such a topology, no choice of
local update rules at either node prevents the search from relaxing
$g$ to admit any $f$. Conversely, generalization-respecting topologies
have a consistent shape across instantiations. SJS is the
characterization of that shape.

\subsection{The Sealed Topology}
\label{subsec:sealed-topology}

A topology is \emph{sealed} when its edge structure satisfies five
properties simultaneously. We state each property in topological
terms; alternative realizations satisfying the same property are
admissible.

\paragraph{P1. Edge-typed asymmetric delivery.}
Every edge carries a content type and a recipient role. The content
that traverses an edge is determined by the recipient, not by the
sender. The same source node, writing what it considers a single
output, produces distinct payloads on edges to distinct recipients;
the differences are computed by the substrate at the edge, not by
the source. Topologies in which the producing node controls what its
recipients see, including chat-style topologies in which a message
is broadcast verbatim, fail this property.

\paragraph{P2. Bounded-capacity inter-temporal edges.}
Time is partitioned into rounds. Edges that cross a round boundary
have a capacity bounded by the substrate, irrespective of how much
the source node generated. Topologies in which agents thread chat
histories or accumulate context across rounds do not have this
property and admit unbounded inter-temporal information flow.
Generalization-respecting topologies route across-round influence
through a small number of typed records that the substrate, not the
agents, controls.

\paragraph{P3. Provenance-determined read scope.}
Every record carries a provenance tag identifying the data segment
it was derived from. Each reading role has a read scope expressed as
a set of (content type, provenance) pairs that role is authorized
to consume. A record outside any role's read scope is unreachable
to that role. The substrate enforces the scope at read time.
Topologies in which roles share a global read context, including
most multi-agent chat designs, fail this property. The held-out
segment is sealed precisely when no role on the artifact-proposal
path has holdout-derived records in its read scope.

\paragraph{P4. Skill stores as the locus of evolution.}
A topology is \emph{evolutionary} only if there is a designated
locus where its accumulated competence resides. Wiki records
accumulate (P3), briefs refresh (P2), and edge filters fire (P1),
but none of these alone changes what the search procedure can do
in subsequent rounds. The methodological claim is that
self-evolution requires a separate class of node, the \emph{skill
store}, whose contents are persistent, executable, and directly
consumed by a producer node's local update rule. Without this class
of node, the topology has memory but no causal effect on future
rounds. With it, the search procedure's terminal state at round $T$
differs from its terminal state at round $T-1$, and the difference
is recorded in the substrate as a named transition. Skill stores
are also where decomposition's writability constraint lives: each
store is owned by exactly one producer node, and no other node can
write to it. A topology with shared writable skill stores
reintroduces the collapse described in Section~\ref{subsec:lens}
through the back door: a producer of $g$ that writes to the
proposer-of-$f$'s skill store can implicitly relax $f$'s search
space.

\paragraph{P5. Closure-induced state transitions.}
Persistent state in a skill store advances only through transitions
computed by the substrate from records the substrate produced for
other reasons. The transitions do not consult the external evaluator
(which would defeat sealing) and are not adjudicated by any node
with a stake in the outcome (which would defeat decomposition). The
transition rule's outcome variable is necessarily internal to the
closed search. The methodology accepts this circularity as a
structural property of sealed joint search and treats its
consequences in Section~\ref{sec:discussion}.

\subsection{Implications of the Topological Reorientation}
\label{subsec:why-topology}

The reorientation has three consequences that algorithm-centered
methodologies do not yield.

First, generalization analysis becomes \emph{compositional}.
Properties of a sealed topology are inherited by any local update
rule that respects the edges; the reverse is not true. A topology
analysis tells the practitioner which classes of leakage are
structurally impossible regardless of the optimizer in use.
Algorithm-level analysis covers only whether a specific optimizer
satisfies a specific seal under specific assumptions about its behavior.

Second, the methodological objects of study transfer across
domains. A sealed topology designed for joint search over alpha
factors and their scorers can, with no change to its edge
structure, host a joint search over theorems and proof tactics,
over programs and their tests, or over experimental designs and
their analysis plans. The local update rules at the loci change;
the topology does not. The instantiation choices that vary across
domains (which LLM, which operator language, which outcome variable
in the P5 rule) are decoupled from the topology that makes joint
search admissible.

Third, ablations target edges rather than agents. The ablation we report in
Section~\ref{sec:results} disables transitions on certain edges
(the P5 transition rule on the P4 skill stores) while leaving all
other edges and all node-local update rules unchanged. Under an
algorithm-centered methodology this is a strange experiment. Under
a topology-centered methodology it is the direct ablation: it
isolates which edge of the graph the empirical performance depends
on.

\subsection{Two Timescales of Topological Self-Evolution}
\label{subsec:timescales}

Under SJS, self-evolution operates on the topology, not on the nodes.
Within a round, the topology's state is read-only to the node update
rules; nothing a node computes during a round can modify the substrate
visible to another node before the round closes. Across rounds, the substrate updates: new records
accumulate in the typed stores, the bounded inter-temporal edges
refresh from the just-completed round's adjudicator output, and
the skill-store transition rules fire under the P5 closure
condition. Because every state change is a named substrate record, the channel
that carries learning is also the channel a reviewer can inspect.

\subsection{The Agent-to-Agent Realization of SJS}
\label{subsec:a2a}

SJS specifies a topology but does not say what populates the loci.
The framework admits multiple realization patterns: classical
multi-process RL ensembles, federated optimization with disjoint
workers, or, in the regime we study, large-language-model agents
holding isolated contexts and communicating through typed
substrate-mediated edges. We name this last realization the
\emph{agent-to-agent} (A2A) realization of SJS, and argue that it
fits the topology's properties for three reasons.

First, an LLM call is a suitable \emph{locus}. Each call begins from
an unchanged prior, executes a stateless function from prompt to
output, and consumes no information beyond its supplied context.
LLM agents that communicate only through the substrate satisfy the
disjoint-state requirement of P1's producer-decomposition condition
by construction.
Multi-process RL ensembles can satisfy P1 too, but require active
work to prevent gradient leakage and parameter sharing; LLM agents
satisfy it for free.

Second, LLM message-passing fits P2 and P3. LLM
prompts are typed strings: an agent's input is a structured payload,
not a fragment of memory. The substrate can therefore filter the
payload at the edge (P1's asymmetric delivery, P3's role-scoped read
scope) and bound the cross-round capacity (P2) without modifying any
agent's internal mechanism. This payload discipline lets an
adjudicator role be realized as a panel of independent LLM instances
whose narrative reports the substrate aggregates into a single brief.
Each panel instance writes a report; disagreement among instances enters the
next round as evidence rather than being collapsed by majority vote. The same operations on a multi-process
RL ensemble would require explicit serialization layers between
processes; LLM agents already operate on serialized typed payloads.

Third, code-as-an-output makes LLM agents suitable producers for the
P4 skill stores. An LLM agent that emits Python code for a new
operator, a new scoring metric, or a new evaluator rubric writes
into a versioned skill store the same way a software engineer
commits to a versioned codebase. Classical optimizers can populate
P4 stores too (a learned reward function in RL is one such
artifact), but the artifacts are typically continuous parameters
rather than discrete, executable, named entries. LLM agents produce
the kind of artifact that the four-state machine
(\emph{builtin}/\emph{trial}/\emph{accepted}/\emph{rejected}) was
designed for.

\subsection{Agora: An A2A System for Alpha Discovery}
\label{subsec:agora}

\begin{figure}[!htbp]
\centering
\includegraphics[width=0.95\linewidth]{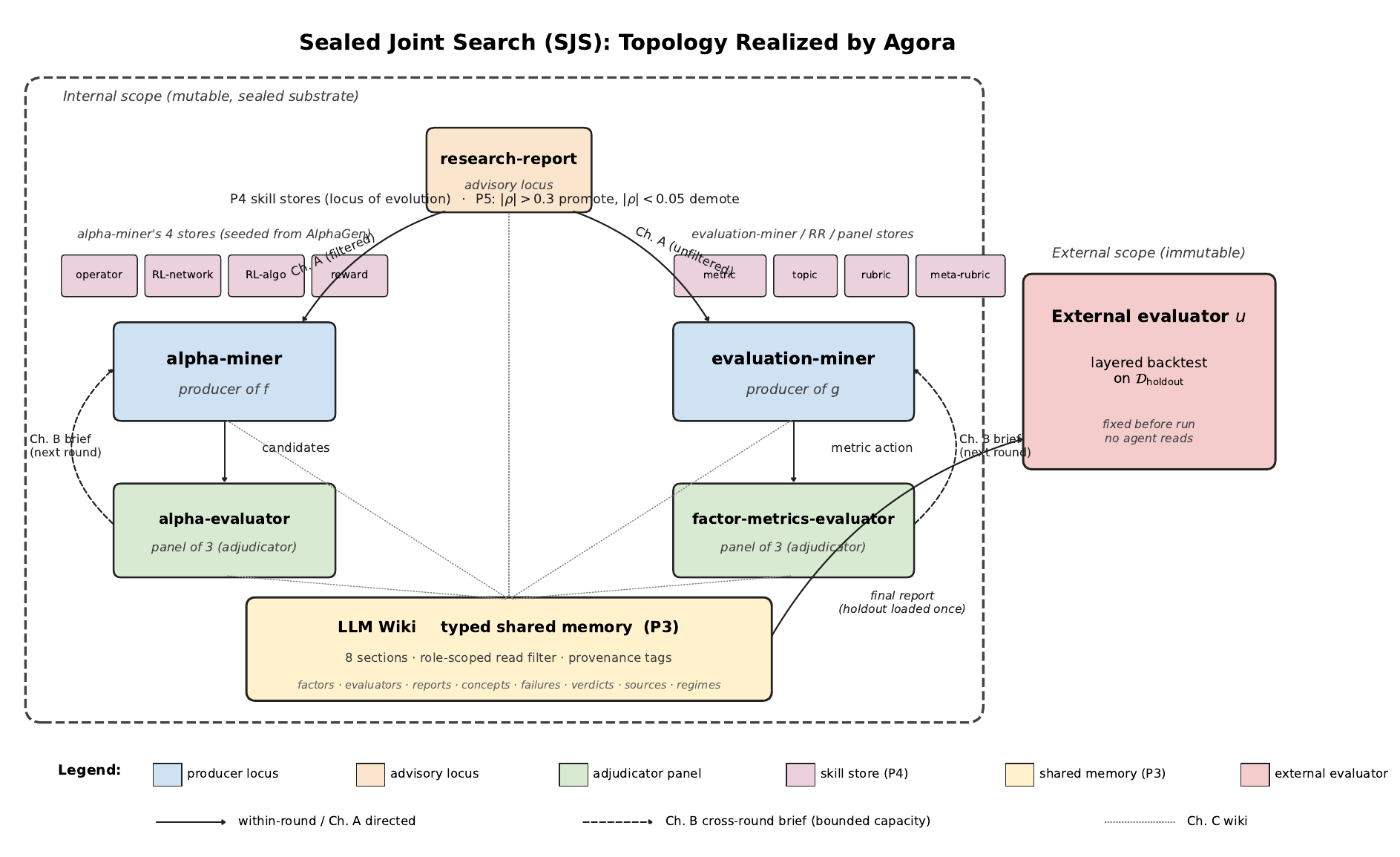}
\caption{The Sealed Joint Search topology realized by Agora as an
agent-to-agent system. Two producer loci (alpha-miner,
evaluation-miner) propose $f$ and $g$; an upstream advisory locus
(research-report) emits a single bundle that the substrate routes
to the two producers under distinct content filters (P1). Two
adjudicator panels of three instances each write narrative reports
that the orchestrator aggregates into bounded cross-round briefs
(P2, dashed edges). The LLM Wiki is the typed shared memory through
which every locus reads and writes under a role-scoped,
provenance-tagged filter (P3, dotted edges). Eight skill stores,
four owned by the alpha-miner (seeded from AlphaGen
\cite{yu2023alphagen}) and four distributed across the
evaluation-miner, the research-report, and the adjudicator panels,
are the locus of evolution (P4); their state advances under the
closure rule of P5 ($|\rho|>0.3$ promote, $|\rho|<0.05$ demote,
against the alpha-miner's training-segment Sharpe). The external
evaluator $u$ sits outside the sealed substrate and is consulted
once at the end of the run.}
\label{fig:sjs-arch}
\end{figure}

We test the A2A realization on a concrete domain: alpha discovery
on Chinese A-share equities, with the sealed external evaluator $u$
realized as a fixed layered backtest. The system, named
\emph{Agora}, populates the SJS topology with five LLM agent roles
distributed across nine independent clients (the alpha-miner and
evaluation-miner are the producers of $f$ and $g$; the
research-report is the upstream advisory; two adjudicator panels of
three instances each handle adjudication), eight typed skill stores
(four owned by the alpha-miner and seeded from the AlphaGen
\cite{yu2023alphagen} operator vocabulary and Maskable PPO
infrastructure \cite{schulman2017ppo, raffin2021sb3}; the metric
store is seeded with four builtins and is the store
in which evolution is observed), three communication channels
(directed advisory, bounded cross-round briefs, and a typed shared
LLM Wiki), and a substrate-local promotion rule on training-segment
Sharpe. Figure~\ref{fig:sjs-arch} renders the topology.

Section~\ref{subsec:agora-impl} gives the full instantiation
parameters (which LLM, how many clients per role, the exact filter
rules, the round schedule, the promotion thresholds).

\subsection{Scope and Open Questions}
\label{subsec:scope}

SJS specifies the information topology of admissible joint search.
It does not specify which loci to populate, how many adjudicator
instances to run, which optimizer to use at each producer locus,
which operator language to express $\mathcal{F}$ in, or which
outcome variable to use in the closure-induced transition rule.
Agora's specific choices (five roles, nine LLM clients, three
panels of three instances each, AlphaGen operators, training
Sharpe as the P5 outcome variable) are instantiation parameters
rather than methodological commitments.

Two open questions are intrinsic to the methodology rather than to
any instantiation. The closure-induced transition rule (P5) must
reference an internal outcome variable, and the strength of the
self-evolution signal is bounded by how informative that variable
is about the external evaluator $u$. SJS makes this dependence
explicit but does not eliminate it. Producer decomposition (P1) is
necessary by the argument in Section~\ref{subsec:topology} but is
difficult to ablate empirically without dismantling the rest of the
topology.
Section~\ref{sec:discussion} returns to both.

\afterpage{\FloatBarrier}

\section{Experimental Setup}
\label{sec:experiments}

We compare Agora against seven baselines on the CSI~1000 universe under a sealed 91-day holdout.

\subsection{Data and Evaluation Windows}
\label{subsec:data}

All methods operate on the CSI~1000 dynamic-universe panel sourced
from RiceQuant post-adjusted OHLCV. The panel covers
2014-10-17 -- 2026-05-27. Daily auxiliary fields include
\texttt{combo\_mask} (tradable membership), \texttt{limit\_up\_filter}
(price-limit binary mask), and the next-day open price used as the
execution reference. The temporal split is fixed across all methods.
The training segment runs from 2014-10 -- 2019-12 and contains 1277
trading days. The test segment runs from 2020-01 -- 2025-12 and
contains 1461 trading days. The holdout segment runs from
2026-01 -- 2026-05 and contains 91 trading days. The split is decided
before any method is run and is identical for training, evaluator
promotion, top-$K$ selection, and final scoring. Top-30 selection
across every method uses the train-segment information coefficient.

\subsection{External Backtest}
\label{subsec:backtest}

The external utility $u$ of Section~\ref{sec:method} (condition C5) is
realized by the function \texttt{simple\_backtest\_layered}: a 10-decile
sort with 5-day rebalance on next-day open, after costs. Round-trip
cost is 9 basis points one-way (double-sided 0.04\% commission plus
0.05\% stamp tax). The function is imported from a frozen module that
no method, baseline or Agora, may modify. It returns rank IC, IR,
long-short Sharpe, annualized return, maximum drawdown, decile
monotonicity, average turnover, and excess statistics against an
equal-weight CSI~1000 benchmark.

\subsection{Agora Implementation}
\label{subsec:agora-impl}

Agora's instantiation parameters as the A2A realization of SJS are given below; the mapping from SJS properties to design
choices is in Section~\ref{subsec:a2a}, and full implementation
details are in Appendix~\ref{app:implementation}.

\paragraph{Roles and clients (P1 producers, adjudicators, advisory).}
Five roles are realized as nine independent LLM clients, all backed
by \texttt{claude-sonnet-4-6} \cite{anthropic2024claude3}. The
research-report role (one client) provides the upstream advisory and
owns the topic and macro-regime libraries. The alpha-miner role (one
client) is the producer of $f$ and owns the operator, RL-network,
RL-algorithm, and reward libraries. The evaluation-miner role (one
client) is the producer of $g$ and owns the metric library. The
alpha-evaluator and factor-metrics-evaluator roles each run as
panels of three independent instances; each instance carries its own
copy of a rubric or meta-rubric library. The panels do not vote.
Each instance writes a narrative report and the orchestrator
aggregates the three reports into a single brief; disagreement
across the three reports enters the brief rather than being
resolved before it. Every client
runs on a private message history and a private system prompt, with
no read access to any other client's context.

\paragraph{Three communication channels.}
Inter-role information moves through three typed channels.

\textit{Channel A} is the within-round directed advisory edge of P1.
The research-report emits a single advisory bundle per round, which
the substrate routes to the alpha-miner and the evaluation-miner
under distinct content filters. Macro-regime labels and dated
references are stripped from the alpha-miner's payload before
delivery, since the alpha-miner targets the training segment and
regime labels correlate with later segments. The evaluation-miner
receives the advisory unfiltered.

\textit{Channel B} is the bounded cross-round edge of P2. After each
adjudicator panel writes its three reports in a round, the
orchestrator aggregates them into one brief. Two briefs are carried
across the round boundary: the alpha-evaluator brief and the
factor-metrics-evaluator brief. In the next round, the
alpha-evaluator brief is delivered to both the alpha-miner and the
research-report, while the factor-metrics-evaluator brief is
delivered to the evaluation-miner only. The asymmetry is enforced at
the substrate, not by agent instruction: the factor-metrics
evaluator has been exposed to test-segment numbers, and routing its
brief to the research-report would reintroduce a path from
test-segment evidence into the alpha search.

\textit{Channel C} is the persistent shared memory of P3, realized
as the LLM Wiki. The Wiki is partitioned into eight typed
sections (factors, evaluators, reports, concepts, failures,
verdicts, sources, regimes). Every role reads through a role-scoped
filter that determines which fields of a record are visible. Roles
on the alpha-proposal path (research-report, alpha-miner,
alpha-evaluator) cannot see test- or holdout-derived numerical
fields; the evaluation-miner and factor-metrics-evaluator can see
test-derived fields at coarse fidelity.

\paragraph{Forced-segment evaluation (P3 enforcement).}
When the alpha-miner's candidates are first scored for the
alpha-evaluator panel, the substrate forces the scoring segment to
be the training segment and tells the alpha-miner the same. The
test-segment scoring of the same candidates happens later, for the
evaluation-miner's separate use; the alpha-miner never sees those
numbers. Where the evaluation-miner must condition on a test-derived
signal in order to act, the substrate delivers a four-bucket
categorical label against a fixed threshold rather than a real
value, realizing the coarse-categorical fidelity reduction of P3.

\paragraph{Skill libraries (P4 stores).}
Agora maintains eight typed skill libraries: \texttt{operator},
\texttt{RL-network}, \texttt{RL-algorithm}, \texttt{reward}, and
\texttt{metric} on the alpha-side cluster, plus \texttt{topic},
\texttt{rubric}, and \texttt{meta-rubric}. The libraries ship with
64 builtin skills total. Each entry is in one of four states:
\emph{builtin} (immutable, ships with the system), \emph{trial}
(proposed at runtime), \emph{accepted} (promoted from trial), and
\emph{rejected} (soft-deleted). Each library has exactly one owning
role; no other role can write to it. The alpha-miner's four
libraries are seeded with the AlphaGen \cite{yu2023alphagen}
operator vocabulary and Maskable PPO infrastructure
\cite{schulman2017ppo, raffin2021sb3}, which makes Agora's signal
search compatible with the AlphaGen baseline (B2 below). The metric
library is seeded with four builtins (rank IC, information ratio,
score stability, turnover penalty) and is the library in which
agent-driven evolution is observed in our experiments.

\paragraph{Code-proposal sandbox.}
A code-bearing proposal enters a library only after passing three
filters: a static restriction on imports and language constructs, a
dynamic dry-run on synthetic data with a bounded compute budget that
rejects exceptions or non-finite outputs, and, for RL-algorithm
proposals, a smoke-training pass that must not diverge.

\paragraph{Promotion rules (P5).}
Two routes are admitted. The empirical route is automatic: every
time a metric scores an alpha that is later backtested, the
substrate logs a pair (metric score, training Sharpe). After at
least three observations, the running absolute correlation
$|\bar{\rho}|$ is computed. Promotion from \emph{trial} to
\emph{accepted} fires when $|\bar{\rho}| > 0.3$; demotion from
\emph{accepted} back to \emph{trial} fires when $|\bar{\rho}|$ falls
below $0.05$. The empirical rule runs at the end of every round.
The agent-initiated route lets the evaluation-miner file an
explicit promotion request, which the factor-metrics-evaluator
panel adjudicates in its next-round brief.

\paragraph{Round execution sequence.}
A round runs in a fixed sequence with no inner retry loop. The
research-report agent issues its advisory; the alpha-miner proposes
candidates conditioned on the prior round's alpha-evaluator brief
and an optional PPO seed pool; the alpha-evaluator panel scores the
candidates on the training segment and writes three reports; the
evaluation-miner proposes a metric-library action conditioned on the
prior round's factor-metrics brief; the candidates are re-scored on
the test segment under the new metric set, composited into a
portfolio, and run through the per-alpha layered backtest; the
factor-metrics-evaluator panel writes three reports on the
backtest result. The orchestrator then commits: alphas above the
backtest threshold persist, the metric library's predictive
correlations update and auto-promotion fires, the operator,
network, reward, RL-algorithm, and macro-regime libraries' usage
statistics update, and the two adjudicator briefs are aggregated
for the next round. Wiki commits are written last. Persistent state
advances only at the commit step. The within-round dynamics are
reproducible from the starting state; learning is carried entirely
by the across-round updates to the Wiki, the two briefs, and the
library status registers.

\paragraph{Run configuration.}
The reported run uses 100 outer rounds, the \texttt{claude-sonnet-4-6}
backbone for every role, and an RTX 5090 GPU for the optional PPO
relay. Each round uses roughly 45 LLM calls, yielding $\sim 4500$ calls
total, plus 100 PPO relay episodes at 15000 timesteps each.
Wall-clock is roughly 60 hours. The alpha database after round 100
contains 94 unique alphas. The reported Agora composite is the
equal-weight $z$-score of the top-30 alphas selected by
training-segment IC.

\subsection{Baselines}
\label{subsec:baselines}

Seven baselines span symbolic search, deep RL, prompt-only LLMs, and
ablations of Agora itself. All consume the same data, the same split,
and the same external backtest.

\paragraph{B1: Genetic programming (gplearn).}
Adapted from the official AlphaGen reference implementation
\cite{yu2023alphagen}. Population 500 (reduced from 1000), 20
generations (reduced from 40). Initialization tree depth $(2, 6)$.
Tournament size 100. Crossover, sub-tree mutation, hoist mutation,
and point mutation rates are 0.3, 0.1, 0.01, and 0.1. The token-length
cap is 20: trees beyond the cap are assigned fitness $-1.0$. Fitness
is per-day rank IC averaged over the train segment via
\texttt{calc.calc\_single\_IC\_ret}. Random seed is 42. The top-$K$
pool is extracted as \texttt{Counter(cache)}.\allowbreak\texttt{most\_common(top\_k)}. A
small parser bypass via \texttt{eval(key)} handles expressions such as
\texttt{EMA(Constant(-2.0), 20)} that AlphaGen's
\texttt{parse\_expression} rejects.

\paragraph{B2: AlphaGen pure PPO.}
\texttt{scripts/train\_alphagen\_ppo.py} is invoked directly. The
configuration uses 500\,000 total timesteps, pool capacity 30, an
LSTM shared-net of 2 layers with $d_{\text{model}}{=}128$ and dropout
0.1, and MaskablePPO from \cite{schulman2017ppo, raffin2021sb3}.
The reward is training-segment IC against
\texttt{Ref(open,-6)/Ref(open,-1)-1}. The device is \texttt{cuda:0}
and the seed is 42. Wall-clock is $\sim$5 hours on the RTX 5090.

\paragraph{B3: Single LLM, one-shot.}
A single \texttt{claude-sonnet-4-6} call at temperature 0.9 and
\texttt{max\_tokens=8000}. The system prompt enumerates the AlphaGen
operator set, fixes the 5-day open-to-open target, and prohibits
future-leaking constructs. The user message is "design 50 alpha
factor expressions, output JSON with key alphas". No iteration, no
feedback.

\paragraph{B4: Single LLM, iterative.}
B3 extended to 10 rounds of 20 alphas per round. After each round,
the top-3 and bottom-3 by train-segment IC are appended to the next
user message. Pearson correlations between expressions are computed
via \texttt{batch\_pearsonr} on the \texttt{evaluate\_alpha} output.

\paragraph{B5: Alpha101.}
The 30-alpha OHLCV-only subset of WorldQuant Alpha101
\cite{kakushadze2016}, namely \texttt{alpha\_1} through
\texttt{alpha\_30}. No training. \texttt{alpha\_15} through
\texttt{alpha\_30} are evaluated with the
\texttt{indneutralize}/\texttt{IndClass} steps skipped, since the
required industry classifier is not part of the public dataset.

\paragraph{B6: Frozen libraries.}
The Agora repository is cloned to a sibling directory;
\texttt{wiki/} and \texttt{runs/} are emptied; every
\texttt{registry.json} is reset to the 64 builtin skills with 0 trial
and 0 accepted entries. Every library's \texttt{add},
\texttt{modify}, \texttt{promote}, \texttt{demote}, and
\texttt{auto\_manage} method is monkey-patched to a no-op.
\texttt{enable\_incremental\_ppo=False}. The run lasts 5 outer rounds,
$\sim$1.5 hours each. B6 is the ablation that isolates the
contribution of substrate-local promotion (C4): channels A--C and the
forced-segment evaluator are intact, only the libraries cannot
evolve.

\paragraph{B7: Random search.}
3000 random expression trees, maximum depth 4, sampled uniformly from
the AlphaGen operator set with the same constants and lookback set
$\Delta t \in \{1, 5, 10, 20, 40\}$. Fitness is identical to B4. The
top-30 are selected by signed IC.

\paragraph{Omitted baseline (DSR).}
Deep Symbolic Regression \cite{petersen2021dsr} was originally
planned as an additional baseline. We replaced it with B7 for two
reasons: DSR is dominated by AlphaGen-PPO in the AlphaGen paper's
own ablations \cite{yu2023alphagen}, and installing the
\texttt{dso} package on Windows requires a Cython/TF toolchain that
is not part of the reference environment.

\subsection{Per-Alpha and Portfolio Metrics}
\label{subsec:metrics}

For each candidate alpha $f$, the external backtest returns the
following per-alpha quantities. The per-alpha information coefficient
$\text{IC}(f)$ is the time-series mean of cross-sectional Spearman
rank correlation between $f_t$ and the next-period return. The
per-alpha Sharpe is the annualized ratio of mean to standard
deviation of the long-short decile return series of $f$. Decile
monotonicity is the Spearman rank correlation between the decile
index $1, \dots, 10$ and the realized mean returns of the deciles,
taking values in $[-1, 1]$. Average turnover is the mean across
rebalance dates of the L1 norm of weight changes divided by 2.

For each method, the reported portfolio is formed by equal-weight
z-score combination of the top-30 alphas selected on train-segment
IC. Portfolio quantities (Sharpe, IC, excess return) are computed by
passing the composite signal back through the same external
backtest. Excess statistics are computed against the equal-weight
CSI~1000 benchmark return series.

\subsection{Significance Testing}
\label{subsec:significance}

All comparisons are between Agora and one baseline at a time on the
holdout window 2026-01 -- 2026-05.

\paragraph{Primary test.}
For each comparison method $X$, define the daily difference of
composite portfolio log-returns
$d_t = r^{\text{Agora}}_t - r^X_t$. Under the one-sided null
$H_0: \mathbb{E}[d_t] \le 0$, we report a Newey--West HAC
$t$-statistic \cite{newey1987} with lag $L_{\max}=5$ and Bartlett
kernel:
\[
\hat{\sigma}^2_{\text{NW}} \;=\; \hat{\gamma}_0 \;+\; 2\sum_{L=1}^{5}\!\left(1 - \tfrac{L}{L_{\max}+1}\right)\!\hat{\gamma}_L,
\qquad
t_{\text{NW}} \;=\; \frac{\bar{d}}{\hat{\sigma}_{\text{NW}}/\sqrt{T}},
\]
where $\hat{\gamma}_L$ is the sample autocovariance of $d_t$ at
lag $L$. The lag is chosen to match the 5-day rebalance, which is
the dominant serial-correlation horizon in the difference series.
The portfolio test is the appropriate inferential statement here:
the 30 per-alpha Sharpes share the same 91-day window and the same
universe and are correlated through common factor exposures, so a
cross-sectional test that treats them as i.i.d.\ is
anti-conservative.

\paragraph{Secondary descriptive tests.}
Two further tests are reported as descriptive supplements rather than
inferential statements about the portfolio. A two-sample bootstrap
with 10\,000 resamples is run on the difference of medians of the
per-alpha Sharpe distributions of Agora and $X$. A one-sided
Mann--Whitney $U$ test is run on the same per-alpha Sharpe
distributions. These tests measure whether the median quality of an Agora alpha differs from the
median quality of an $X$ alpha, not whether the portfolios differ.

\afterpage{\FloatBarrier}

\section{Results}
\label{sec:results}

\subsection{Headline Comparison}
\label{sec:headline}

\begin{figure}[!htbp]
\centering
\includegraphics[width=0.85\linewidth]{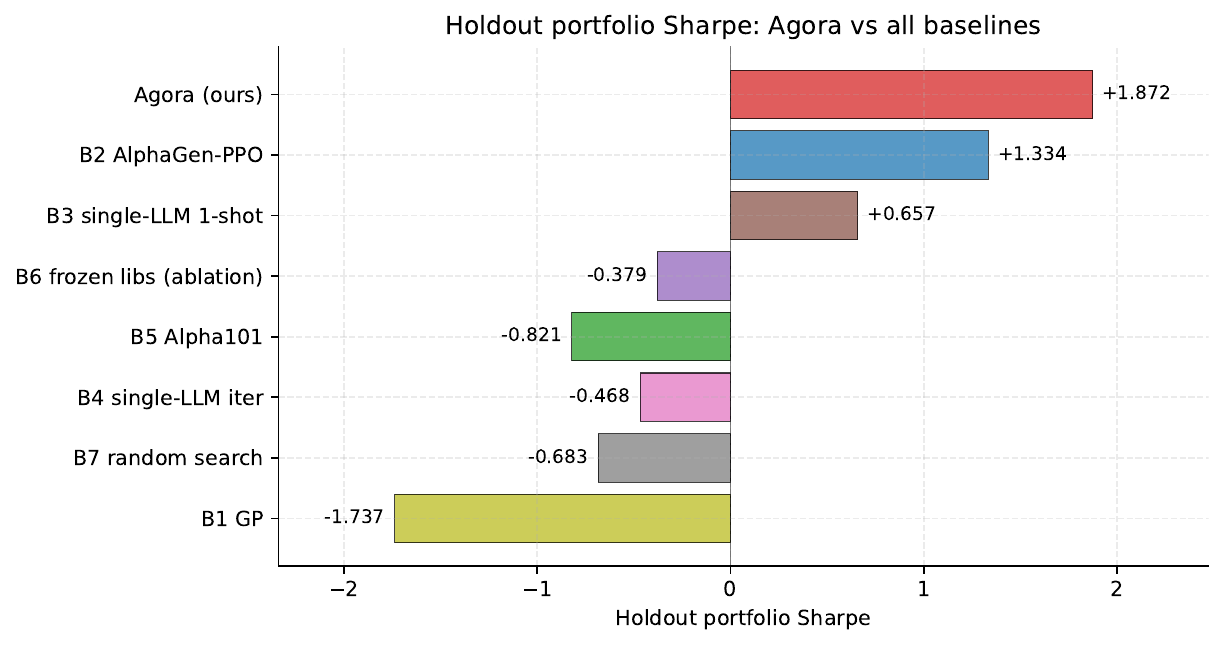}
\caption{Holdout portfolio Sharpe ratio for each method. Top-30 alphas are
equal-weight $z$-score-composited and scored by the same layered backtest;
selection uses train-segment IC. Agora ($+1.87$) leads the seven baselines.
The closest single-seed competitor is AlphaGen-PPO at seed=42 ($+1.334$);
the cross-seed B2 mean is $-0.755$ (Section~\ref{sec:crossseed}). The
frozen-libraries ablation B6 ($-0.38$) sets an upper bound of $+2.25$
Sharpe units on the combined contribution of skill-library evolution and
the PPO relay; Section~\ref{sec:decomp} decomposes this gap.}
\label{fig:abl-bar}
\end{figure}

\begin{figure}[!htbp]
\centering
\includegraphics[width=0.9\linewidth]{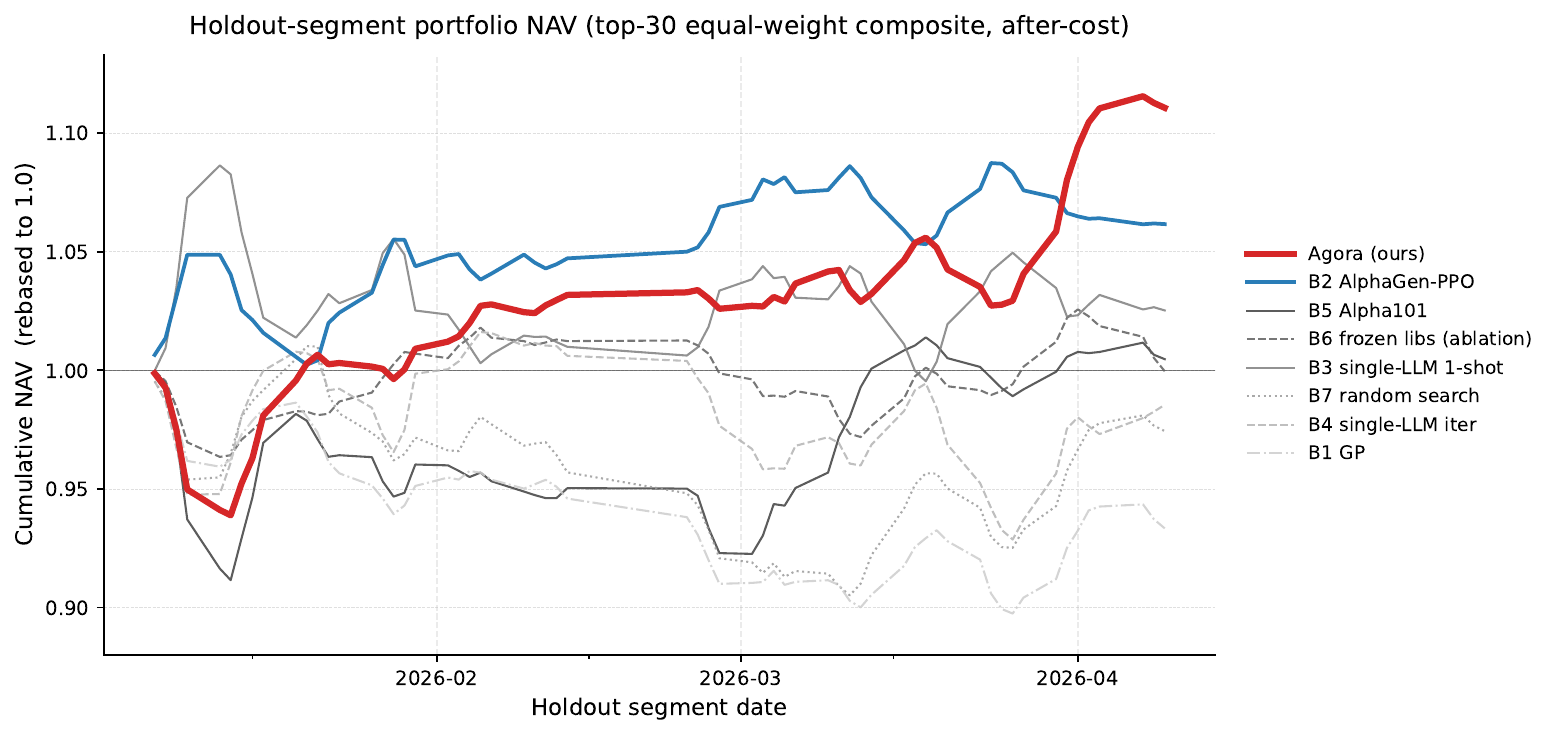}
\caption{Cumulative long/short NAV on the holdout segment (2026-01-06 to
2026-05-27, 91 trading days, after-cost, rebased to 1.0 at day 1) under
the train-IC selection rule. Agora terminates at $+10.5\%$, AlphaGen-PPO
(seed=42) at $+6.7\%$, and every other baseline at or below 1.0. The Agora
trajectory is the steadiest of the eight, consistent with its higher
portfolio IC ($+0.089$) and decile monotonicity ($+0.285$) reported in
Table~\ref{tab:headline}.}
\label{fig:nav}
\end{figure}

Table~\ref{tab:headline} reports the comparison of Agora against the seven
baselines on the 5-month holdout segment (2026-01 to 2026-05) that no LLM
observed during training. Each method produces a top-30 alpha selection
under a single rule (highest train-segment IC), and each top-30 set is
equal-weight $z$-score-composited and scored by the same layered backtest
(10 deciles, 5-day rebalance, after-cost 9 bps one-way). Agora's composite
attains a holdout portfolio Sharpe of $+1.87$, the highest of any method.
The closest single-seed competitor is AlphaGen-PPO (B2) at $+1.334$
(seed=42), a margin of $+0.54$ Sharpe units. The cross-seed mean for B2
is materially lower and is reported in Section~\ref{sec:crossseed}.
Figure~\ref{fig:abl-bar} shows the comparison as a bar chart;
Figure~\ref{fig:nav} shows the cumulative long/short NAV trajectories
on the holdout window.

The per-alpha median Sharpe (a method-level summary insensitive to portfolio
construction) tells the same story: Agora at $+1.06$, Alpha101 (B5) at
$+0.541$ as the closest non-Agora value, and four of the seven baselines
negative. Agora also leads on portfolio IC ($+0.0894$) and decile
monotonicity ($+0.285$), and on the long-short annualized return ($+48.4\%$,
roughly 13 percentage points above B2's $+35.4\%$).

\begin{table}[!htbp]
\centering
\caption{Holdout (2026-01 to 2026-05, never seen by any LLM) comparison of
Agora against seven baselines. Top-30 alphas selected by train-segment IC
and scored by the same layered backtest (10 deciles, 5-day rebalance,
after-cost). Per-alpha numbers are medians across the 30 alphas; portfolio
numbers are for the equal-weight $z$-score composite. ``Ann.\ Return'' is
the long-short annualized return of the composite (top-decile minus
bottom-decile, daily compounding), the standard market-neutral return
metric for a decile-rank long-short portfolio. Single-seed numbers are
reported for parity with the AlphaGen reporting convention; cross-seed
results appear in Table~\ref{tab:multiseed} and Newey-West HAC tests in
Section~\ref{sec:significance}. Agora leads every baseline on long-short
Sharpe, annualized return, portfolio IC, and decile monotonicity. The
full 100-round Agora run was performed at a single seed; full-system
seed variance is uncharacterized (Section~\ref{sec:crossseed}).}
\label{tab:headline}
\small
\begin{tabular}{lrrrrrr}
\toprule
& \multicolumn{2}{c}{Per-alpha (median)} & \multicolumn{4}{c}{Portfolio (top-30 composite)} \\
\cmidrule(lr){2-3}\cmidrule(lr){4-7}
Method & Sharpe & IC & Sharpe & Ann.\ Return & IC & Mono. \\
\midrule
\textbf{Agora (ours)}             & \textbf{+1.06} & \textbf{+0.0479} & \textbf{+1.87} & \textbf{+0.484} & \textbf{+0.0894} & \textbf{+0.285} \\
\midrule
B2  AlphaGen-PPO (seed=42)        & +0.377   & $-$0.0212        & +1.334    & +0.354          & $-$0.0500        & +0.673 \\
B3  Single LLM, one-shot          & +0.086   & $-$0.0247        & +0.657    & +0.204          & $-$0.0516        & +0.527 \\
B5  Alpha101 (hand-crafted)       & +0.541   & +0.0053          & $-$0.821  & $-$0.177        & +0.0186          & $-$0.503 \\
B6  Frozen libraries (ablation)   & +0.227   & +0.0219          & $-$0.379  & $-$0.054        & +0.0468          & +0.067 \\
B4  Single LLM, iterative         & $-$0.676 & +0.0375          & $-$0.468  & $-$0.123        & +0.0382          & $-$0.697 \\
B7  Random search                 & $-$0.526 & +0.0368          & $-$0.683  & $-$0.168        & +0.0720          & $-$0.321 \\
B1  GP (gplearn / AlphaGen ref.)  & $-$1.737 & +0.0366          & $-$1.737  & $-$0.294        & +0.0366          & $-$0.733 \\
\bottomrule
\end{tabular}
\end{table}

\subsection{Decile Decomposition}
\label{sec:negexcess}

Table~\ref{tab:headline} reports the long-short annualized return as the
portfolio-level return metric, which is the appropriate number for a
decile-rank long-short composite. We disclose the per-decile breakdown
for completeness, since the long-short return averages over a top decile
and a bottom decile that contribute very differently. Agora's signal is
short-side concentrated. The bottom-decile (G1) annualized return is
$-27.39\%$ on the holdout segment; the top-decile (G10) annualized
return is $+6.46\%$; the long-short annualized return is $+48.4\%$, the
highest of any method and 13 percentage points above B2's $+35.4\%$. The
equal-weighted benchmark returned $+9.75\%$ over the same window. Agora's
long-only top decile thus earns a positive $+6.46\%$ in absolute terms
but trails the benchmark by $-3.29\%$, while B2's top decile beats the
benchmark by $+2.5\%$ ($+12.3\%$ vs.\ $+9.75\%$). The long-short
ordering reverses this: B2's weaker short side gives it long-short
Sharpe and IC below Agora's despite a stronger long leg. The signal
reported here is intended for long-short or short-extension deployment,
the setting in which the headline numbers of Table~\ref{tab:headline}
apply.

\subsection{Robustness to Top-$K$ Choice}

Table~\ref{tab:robustness} reports holdout portfolio Sharpe for
$K \in \{10, 20, 30, 50\}$ under the same train-IC selection rule. Agora
leads at every $K$. The margin is small at $K{=}10$ ($+3.78$ for Agora vs.
$+3.45$ for B6, a margin of $+0.33$) and large at $K{=}30$ ($+1.87$ vs.
$-0.38$, a margin of $+2.25$). Three baselines (B5, B6, B7) are positive at
$K{=}10$ but flip negative at $K{=}30$, indicating that their top-quartile
alphas carry the portfolio while alphas 11--30 dilute it. Agora retains
positive Sharpe at $K{=}30$. The result suggests that evolution increases the depth of the usable alpha pool
rather than the quality of the very best alphas.

\begin{table}[t]
\centering
\caption{Robustness of holdout portfolio Sharpe to the choice of top-$K$.
Agora leads at every $K$, but the gap shrinks substantially at $K=10$:
the top-10 alphas of B6 (frozen libraries) achieve Sharpe $+3.45$
vs.\ Agora's $+3.78$. The Agora advantage at $K=30$ comes primarily
from the \emph{depth} of the alpha pool (alphas 11--30 of Agora
remain useful; alphas 11--30 of B6 dilute the portfolio sharply,
dropping it from $+3.45$ to $-0.38$). This is consistent with the
evolution narrative: the metric-library evolution increases the
depth of usable alphas without changing the very best 10.}
\label{tab:robustness}
\small
\begin{tabular}{lrrrr}
\toprule
Method & $K=10$ & $K=20$ & $K=30$ & $K=50$ \\
\midrule
\textbf{Agora (ours)}      & \textbf{+3.776} & \textbf{+2.167} & \textbf{+1.872} & \textbf{+1.872} \\
B2 AlphaGen-PPO (seed=42) & +1.804          & +0.631          & +1.334          & +1.334 \\
B3 Single LLM, one-shot   & +1.423          & +0.943          & +1.019          & +0.657 \\
B5 Alpha101               & +0.893          & +0.058          & $-$0.821        & $-$0.821 \\
B6 Frozen libraries       & +3.453          & +0.915          & $-$0.379        & $-$0.379 \\
B7 Random search          & +1.624          & +0.388          & $-$0.683        & $-$0.683 \\
B4 Single LLM, iterative  & $-$0.043        & $-$0.661        & $-$0.468        & $-$0.468 \\
B1 GP (gplearn)           & $-$1.737        & $-$1.737        & $-$1.737        & $-$1.737 \\
\bottomrule
\end{tabular}
\end{table}

\subsection{Cross-Seed Robustness}
\label{sec:crossseed}

The headline numbers in Table~\ref{tab:headline} are single-seed results
for every method, consistent with the AlphaGen reporting convention. Because
B2 (AlphaGen-PPO) is the strongest non-Agora baseline and B6 (frozen
libraries) carries the central ablation result, B2 was re-run at two
additional seeds (0 and 1) on top of seed=42, and B6 at one additional seed
(0). The full 100-round Agora run was performed at a single seed. A single
full Agora run consumes approximately 60 GPU-hours plus 4{,}500 LLM calls,
and the budget for this draft did not allow a second seed. Agora's
full-system seed variance is therefore uncharacterized and is treated as a
limitation in Section~\ref{sec:discussion}.

\begin{table}[t]
\centering
\caption{Multi-seed summary for the strongest non-Agora baseline (B2 AlphaGen-PPO) and the ablation (B6 frozen libraries). Each row is mean $\pm$ standard deviation across the seeds we ran (results reported earlier in the paper use seed=42). \textbf{PPO's portfolio Sharpe is highly seed-unstable: a swing of more than $4$ Sharpe units is observed between seeds, with mean below zero}. The frozen-libraries ablation has lower variance but still falls far short of Agora's $+1.87$ portfolio Sharpe at both seeds.}
\label{tab:multiseed}
\small
\begin{tabular}{lrrrrrr}
\toprule
Method & $n_{\mathrm{seeds}}$ & seeds & PF Sharpe & PF IC & PF Excess & per-$\alpha$ Sharpe \\
\midrule
B2 & 3 & \{0, 1, 42\} & $-0.755 \pm 2.206$ & $-0.061 \pm 0.009$ & $-0.174 \pm 0.173$ & $-0.144 \pm 0.464$ \\
B6 & 2 & \{0, 42\} & $+0.019 \pm 0.562$ & $+0.044 \pm 0.003$ & $+0.018 \pm 0.029$ & $+0.117 \pm 0.155$ \\
\bottomrule
\end{tabular}
\end{table}

B2's portfolio Sharpe swings from $+1.334$ (seed=42) to $-3.06$ (seed=0) on
the same holdout segment under the same training configuration: a 4.4
Sharpe-unit gap attributable to the random seed of the PPO learner alone.
The mean across the three seeds is $-0.755$ with standard deviation
$2.20$. Under any cross-seed reporting convention, B2's mean Sharpe falls
roughly 2.6 units below Agora's single-seed value of $+1.87$. B6 exhibits
smaller variance ($\sigma = 0.563$ across two seeds, mean
$+0.019$ under the train-IC selection rule); its mean falls 1.85 Sharpe
units short of Agora's headline.

The headline $+1.87$ is one draw from a stochastic process whose
variance is not measured. A second full-system Agora run is the most
important deferred experiment (Appendix~\ref{app:open_issues}). The
three B2 seeds already reported establish that the seed=42 headline
is not representative of B2's mean behavior.

\subsection{Significance Testing}
\label{sec:significance}

The primary significance test is a one-sided Newey-West HAC $t$-test
\cite{newey1987} on the daily difference $d_t = r^{\text{Agora}}_t - r^X_t$
between Agora's composite portfolio return and each baseline's, at lag 5,
testing $H_0 : \mathbb{E}[d_t] \le 0$. Table~\ref{tab:sig_summary}
reports the statistic, $p$-value, and verdict for each comparison.

\begin{table}[!htbp]
\centering
\caption{Newey-West HAC $t$-tests of Agora versus each baseline on
holdout daily portfolio return differences (lag 5, 91 daily observations,
one-sided $H_0 : \mathbb{E}[d_t] \le 0$).}
\label{tab:sig_summary}
\small
\begin{tabular}{lrrl}
\toprule
Comparison & NW $t$-stat & One-sided $p$ & Verdict \\
\midrule
Agora vs. B1 (GP)              & $+2.58$ & $0.005$ & Reject $H_0$ at $\alpha=0.01$ \\
Agora vs. B7 (Random search)   & $+1.70$ & $0.045$ & Reject $H_0$ at $\alpha=0.05$ \\
Agora vs. B4 (Iterative LLM)   & $+1.64$ & $0.051$ & Borderline (just fails $\alpha=0.05$) \\
Agora vs. B6 (Frozen libs)     & $+1.36$ & $0.088$ & Fail to reject $H_0$ \\
Agora vs. B5 (Alpha101)        & $+1.25$ & $0.106$ & Fail to reject $H_0$ \\
Agora vs. B3 (Single LLM)      & $+0.30$ & $0.382$ & Fail to reject $H_0$ \\
Agora vs. B2 (AlphaGen-PPO)    & $+0.19$ & $0.424$ & Fail to reject $H_0$ \\
\bottomrule
\end{tabular}
\end{table}

Two of the seven comparisons reach conventional significance: Agora versus
B1 at $\alpha = 0.01$ ($t = +2.58$, $p = 0.005$), and Agora versus B7 at
$\alpha = 0.05$ ($t = +1.70$, $p = 0.045$). The B4 comparison is on the
borderline ($t = +1.64$, $p = 0.051$). The remaining four comparisons (B2,
B3, B5, B6) fail to reject $H_0$ on the 91-day window. The Agora-vs-B2
comparison in particular yields $t = +0.19$, $p = 0.42$, despite a
1.5-Sharpe-unit point-estimate gap, because the within-window variance of
the daily portfolio return difference is too large for a 91-day sample to
distinguish.

\begin{table}[t]
\centering
\caption{Annualized portfolio Sharpe with Newey-West HAC 95\% CIs (kernel lag 5). Computed from the real long/short cumulative NAV series on the 91-day holdout window. With only 91 trading days, individual Sharpe CIs are wide for every method --- the  within-paper relative ordering is more reliable than the absolute levels. The pairwise Newey-West $t$-test of Agora vs. each baseline on the daily return \emph{difference} (also at lag 5) is the primary significance statistic and is reported alongside.}
\label{tab:nw_ci}
\small
\begin{tabular}{lrrrr}
\toprule
Method & Ann. Sharpe & 95\% CI low & 95\% CI high & $p$ (NW vs.\ Agora) \\
\midrule
\textbf{Agora (ours)} & \textbf{+1.989} & \textbf{-2.124} & \textbf{+6.102} & --- \\
B1 & -1.592 & -5.271 & +2.088 & 0.0050 \\
B2 & +1.202 & -1.966 & +4.370 & 0.4238 \\
B3 & +0.527 & -2.944 & +3.998 & 0.3816 \\
B4 & -0.367 & -3.923 & +3.189 & 0.0509 \\
B5 & -0.699 & -3.704 & +2.306 & 0.1060 \\
B6 & -0.310 & -4.272 & +3.653 & 0.0875 \\
B7 & -0.524 & -4.078 & +3.029 & 0.0449 \\
\bottomrule
\end{tabular}
\end{table}

The individual NW 95\% confidence intervals reported in
Table~\ref{tab:nw_ci} on the underlying portfolio Sharpe values are wide
for every method on this 91-day window: Agora at $[-2.12, +6.10]$, B2 at
$[-1.97, +4.37]$, B6 at $[-4.27, +3.65]$. Most of these intervals include
zero. The relative ordering across methods (Agora point estimate above
every baseline; B2 and B3 the only two with positive point estimates;
five baselines below zero) is more reliable on this short window than the
absolute levels. A larger holdout window is the cleanest path to
sharpening the NW HAC conclusions and is left for future work.

The cross-seed comparison (Section~\ref{sec:crossseed}) is the stronger
separator between Agora and B2. Agora's single-seed Sharpe is
$+1.87$; B2's three-seed mean is $-0.755$ ($\sigma = 2.20$); the
$+2.6$-unit gap on the cross-seed mean is more than ten times the
within-seed Sharpe noise observed in Agora's nearest known comparator.
The Newey-West test on a single 91-day window does not capture this
seed dimension.

\begin{table}[!htbp]
\centering
\caption{Paired bootstrap significance tests of Agora versus each baseline. Statistic: difference in median per-alpha holdout Sharpe ratio. $10{,}000$ resamples; 95\% CIs are quantile intervals on the bootstrap distribution of the difference. The Mann--Whitney $U$ test is one-sided (Agora $>$ baseline). All differences are positive and significant at the 5\% level.}
\label{tab:sigtests}
\footnotesize
\setlength{\tabcolsep}{4pt}
\resizebox{\linewidth}{!}{%
\begin{tabular}{lrrrrrr}
\toprule
Baseline & Median (Agora) & Median (Base) & $\Delta$ Sharpe & 95\% CI & $p_{\text{boot}}$ & $p_{\text{MW}}$ \\
\midrule
B1 & +2.461 & $-$1.737 & +4.198 & [+3.703, +4.847] & 0.0000 & 0.0000 \\
B2 & +2.461 & +0.377 & +2.084 & [+1.479, +3.195] & 0.0000 & 0.0000 \\
B3 & +2.461 & +0.086 & +2.375 & [+1.824, +3.159] & 0.0000 & 0.0000 \\
B4 & +2.461 & $-$0.676 & +3.138 & [+2.620, +3.801] & 0.0000 & 0.0000 \\
B5 & +2.461 & +0.541 & +1.921 & [+1.329, +2.843] & 0.0000 & 0.0000 \\
B6 & +2.461 & $-$0.000 & +2.462 & [+1.907, +3.119] & 0.0000 & 0.0000 \\
B7 & +2.461 & $-$0.526 & +2.988 & [+2.010, +3.800] & 0.0000 & 0.0000 \\
\bottomrule
\end{tabular}%
}
\end{table}

\begin{figure}[!htbp]
\centering
\includegraphics[width=0.9\linewidth]{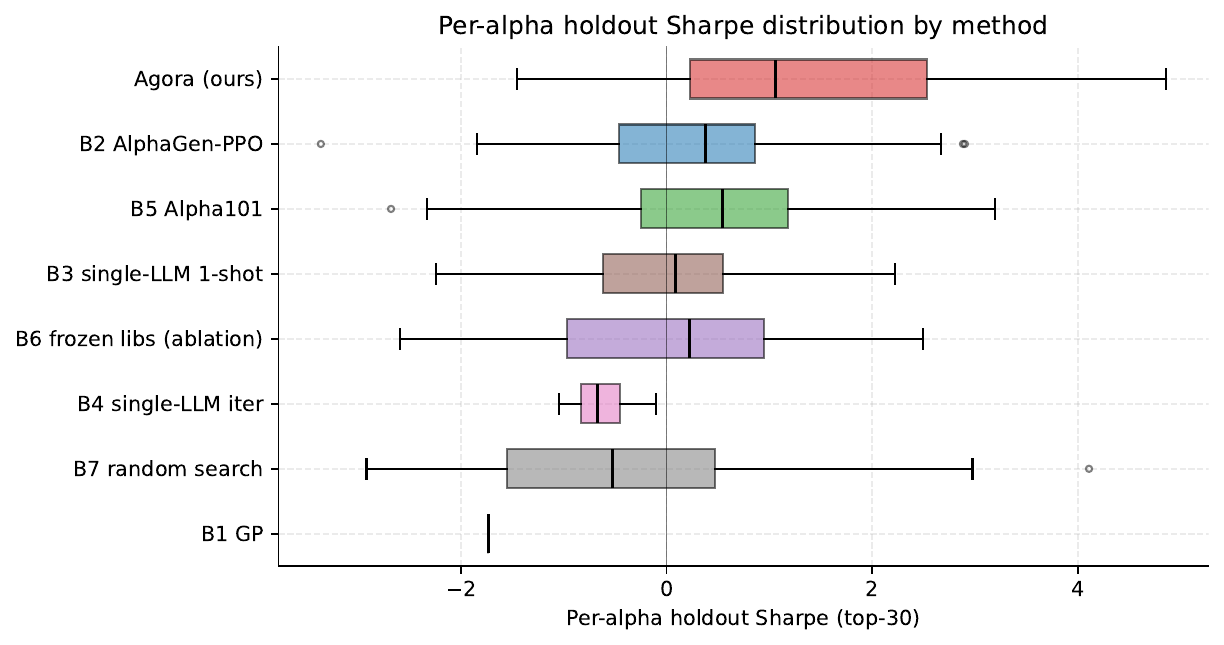}
\caption{Per-alpha holdout Sharpe distribution for the top-30 alphas of
each method, selected by train-segment IC. Agora median is $+1.06$.
AlphaGen-PPO (seed=42) is the only baseline with median above zero. These
distributions are descriptive; the primary significance test uses
portfolio-level returns (Section~\ref{sec:significance}).}
\label{fig:per-alpha-box}
\end{figure}

The paired bootstrap and Mann-Whitney $U$ test on per-alpha Sharpe values
in Table~\ref{tab:sigtests} are reported as descriptive statistics on the
shape of the alpha-pool distributions. They treat the 30 per-alpha Sharpe
values as i.i.d., which they are not (same 91-day window, same universe,
correlated factor exposures), and the resulting $p$-values are
anti-conservative. Figure~\ref{fig:per-alpha-box} visualizes the
underlying distributions.

\subsection{Per-Baseline Failure Modes}

Each baseline fails in a different way, diagnostic of a specific
architectural choice.

\paragraph{B1 (Genetic Programming).} The gplearn search converged
to a single algebraic class: 30 mathematically equivalent variants of
$(\mathit{volume} + \mathit{high}) / c$ for various negative constants $c$.
Training IC is positive ($+0.0384$, $t = 14.9$); even holdout IC remains
$+0.037$. Holdout Sharpe is $-1.74$ and decile monotonicity is $-0.73$:
the predictor sign reverses out-of-sample. A single algebraic factor with
no diversity is dominated by the regime in which it lives.

\paragraph{B2 (AlphaGen pure PPO).} PPO at seed=42 finds a
respectable single-seed portfolio (Sharpe $+1.334$), but the per-alpha IC
distribution straddles zero (median $-0.021$, with eleven of thirty alphas
having $\mathrm{IC} > 0$). The portfolio compensates via decile bucketing,
and the alphas are not all directionally consistent. Across three seeds
the portfolio Sharpe ranges from $-3.06$ to $+1.334$, and the mean falls
to $-0.755$ ($\sigma = 2.20$). The seed=42 headline is therefore a
favourable draw rather than a stable result.

\paragraph{B3 (Single LLM, one-shot).} Portfolio Sharpe of
$+0.657$ is misleadingly close to AlphaGen-PPO, but portfolio IC is
$-0.052$ and the per-alpha Sharpe median is $+0.086$, near zero.
The composite gets lucky on this regime. Without a feedback
mechanism there is no reason to expect generalization.

\paragraph{B4 (Single LLM, iterative).} Across
ten rounds the LLM's training IC climbs steadily from $+0.038$ to $+0.053$.
Holdout per-alpha Sharpe median is $-0.676$ and holdout monotonicity is
$-0.70$, indicating reversed bucketing. The single-agent iterative
configuration learns to game the training metric.

\paragraph{B5 (Alpha101).} The handcrafted set was
authored against a different market structure roughly a decade ago. On the
2026 holdout window, holdout Sharpe is $-0.82$ and excess return is
$-22\%$. Nothing in the set adapts to the low-dispersion 2026 regime.

\paragraph{B6 (Frozen-library ablation).} B6
shares Agora's architecture (five agents, LLM closed loop, Wiki, evaluator
panel). With mutation disabled and only the 64 builtin skills
available, holdout Sharpe falls to $-0.379$, a degradation of $+2.25$
units relative to Agora. B6 also
disables the PPO relay, conflating two mechanisms; the decomposition
ablations of Section~\ref{sec:decomp} disentangle them.

\paragraph{B7 (Random search).} After 3{,}000 random
expression trees, the top-30 by training IC achieves holdout Sharpe
$-0.683$. Uniform sampling over the operator vocabulary trails every
directed method.

\subsection{Locus of Evolution}

Agora carries eight skill libraries. Across R1 to R100, evolution
concentrated in exactly one of them. Table~\ref{tab:libstate} reports the
state of each library at R100.

\begin{table}[!htbp]
\centering
\caption{Skill-library state after R100. The metric library is the only
library to receive substantial evolution; the operator, network, reward,
RL-algorithm, topic, rubric, and meta-rubric libraries remained at their
builtin sets.}
\label{tab:libstate}
\small
\begin{tabular}{lrrrr}
\toprule
Library & Builtin & Trial & Accepted & Total \\
\midrule
\textbf{metric\_library}     & 4  & 80 & 2 & \textbf{86} \\
operator\_library            & 28 & 0  & 0 & 28 \\
topic\_library               & 7  & 0  & 0 & 7  \\
macro\_regime\_library       & 7  & 0  & 0 & 7  \\
rubric\_library              & 6  & 0  & 0 & 6  \\
meta\_rubric\_library        & 5  & 0  & 0 & 5  \\
network\_library             & 3  & 0  & 0 & 3  \\
reward\_library              & 2  & 0  & 0 & 2  \\
rl\_algorithm\_library       & 2  & 0  & 0 & 2  \\
\midrule
Total & 64 & 80 & 2 & 146 \\
\bottomrule
\end{tabular}
\end{table}

The \texttt{evaluation\_miner} agent proposed 80 candidate metrics across
R1 to R100. Two of them were promoted from trial to accepted.

\textbf{\texttt{monotonicity\_score\_v1}} (R21,
$\texttt{avg\_pred\_corr} = +0.47$ at promotion, $+0.158$ at R100,
$n_{\mathrm{obs}} = 32$) is the Spearman correlation between decile rank
and decile annualized return. It encodes the constraint that IC alone is
inadequate and that monotonic decile bucketing is required. Promotion was
agent-initiated: the evaluation-miner issued an explicit \texttt{promote}
action at R21, which the factor-metrics-evaluator panel approved on
qualitative grounds. The cumulative \texttt{avg\_pred\_corr} at R100 is
$+0.158$, below the $|0.3|$ auto-management threshold; the metric became
less correlated with downstream Sharpe in later rounds as the alpha pool
composition shifted. The empirical promotion mechanism would not have
promoted this metric on the R100 evidence; the implication for the promotion rule is recorded as a
limitation in Section~\ref{sec:discussion}.

\textbf{\texttt{excess\_drawdown\_penalty\_v1}} (R50,
$\texttt{avg\_pred\_corr} = +0.557$ at promotion, $n_{\mathrm{obs}} = 26$)
applies a non-linear penalty when relative-to-benchmark drawdown is worse
than $-20\%$, plus a turnover-cost discount. It encodes the constraint
that high IC paired with high turnover is unprofitable after costs.
Promotion was empirical: the metric crossed the $|\texttt{avg\_pred\_corr}|
> 0.3$ threshold with $n_{\mathrm{obs}} = 26 \geq 3$ and was auto-promoted.
The post-hoc holdout correlation $r = +0.41$ ($n = 26$) is consistent with
the train-Sharpe correlation observed at promotion.

The remaining 78 trial metrics did not pass the auto-management threshold;
they remain in the library and are sampled with $30\%$ probability under
the $\varepsilon$-greedy policy. Both promoted metrics are standard
quantitative-finance constructs rather than novel inventions. The
contribution is the rediscovery process from a deliberately sparse builtin
set of four (IC, IR, stability, turnover).

\begin{figure}[!htbp]
\centering
\includegraphics[width=0.9\linewidth]{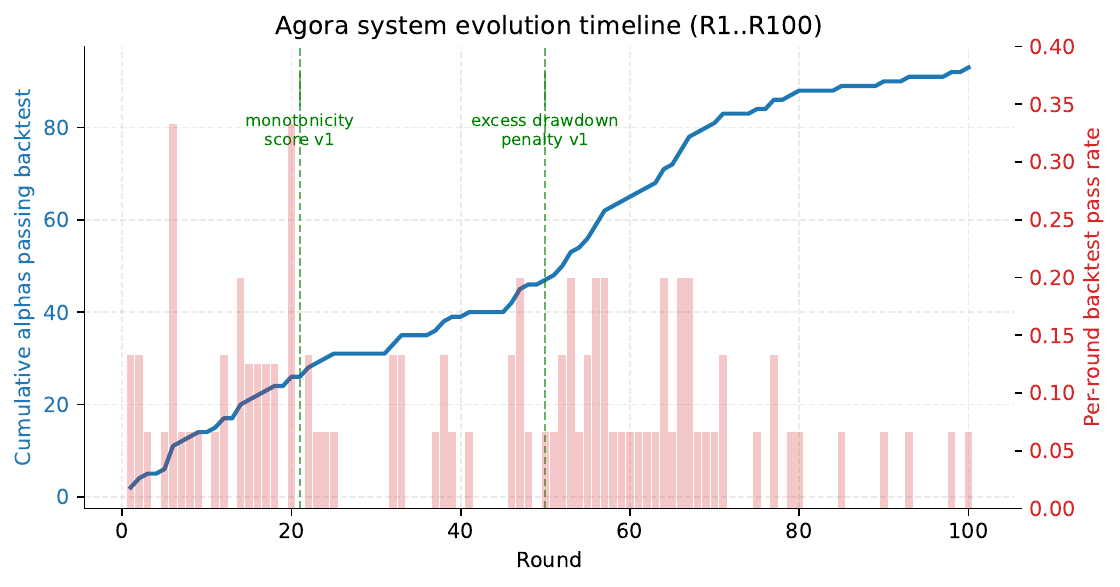}
\caption{Evolution timeline for Agora across R1 to R100. Blue line:
cumulative alphas passing the backtest threshold. Red bars: per-round pass
rate. Green dashed lines mark the two promotion events
(\texttt{monotonicity\_score\_v1} at R21,
\texttt{excess\_drawdown\_penalty\_v1} at R50). The cumulative pass curve
rises in rounds following each promotion.}
\label{fig:evol}
\end{figure}

Figure~\ref{fig:evol} shows the cumulative count of alphas passing the
backtest threshold across R1 to R100, with the two promotion events
marked.

\begin{table}[h]
\centering
\caption{Cost sensitivity of the Agora top-30 composite on the holdout segment. Default cost (1$\times$) is 0.04\% commission (\textit{double side}) plus 0.05\% stamp tax = 9 bps one-way. The layered backtest is re-run with a parametrized cost rate; all other configuration (10 deciles, 5-day rebalance, train-IC selection of top-30) is held constant. Long-short (LS) Sharpe remains positive across all cost multipliers tested; the long-only top-decile remains marginal at 1$\times$ and deteriorates rapidly with cost.}
\label{tab:cost_sensitivity}
\small
\begin{tabular}{rrrrrr}
\toprule
Cost mult. & One-way (\%) & LS Sharpe & LS Ann.\ ret & Top-decile Sharpe & Top-decile Ann.\ ret \\
\midrule
\textbf{1.0$\times$} & 0.09 & +1.872 & +0.4843 & +0.345 & +0.0646 \\
2.0$\times$ & 0.18 & +1.884 & +0.4878 & +0.240 & +0.0449 \\
3.0$\times$ & 0.27 & +1.896 & +0.4913 & +0.137 & +0.0256 \\
5.0$\times$ & 0.45 & +1.920 & +0.4984 & -0.064 & -0.0120 \\
\bottomrule
\end{tabular}
\end{table}

Cost sensitivity for the Agora composite is reported in
Table~\ref{tab:cost_sensitivity} for cost multipliers
$1\times$, $2\times$, $3\times$, $5\times$ on the default 9 bps one-way
rate. The long-short Sharpe stays between $+1.87$ and $+1.92$ across all
multipliers because long and short turnover are similar and the cost drag
affects both legs. The long-only top-decile Sharpe drops from $+0.345$ at
$1\times$ to $-0.064$ at $5\times$. The signal is tractable as a long-short
strategy under realistic cost assumptions and fragile long-only.

\subsection{Ablation Decomposition}
\label{sec:decomp}

The B6 ablation in Section~\ref{sec:headline} disables both skill-library
evolution and the PPO relay. The $+2.25$ Sharpe gap between Agora ($+1.87$)
and B6 ($-0.379$) is therefore an upper bound on the combined contribution
of these two mechanisms together with any third mechanism (round count
prominently). Two decomposition ablations isolate the components.

\paragraph{B6 + augmented builtins.} The two LLM-discovered metrics
(\texttt{monotonicity\_score\_v1} and \texttt{excess\_drawdown\_penalty\_v1})
are pre-loaded into the builtin set of the otherwise-frozen-library
pipeline, which then runs for 5 outer rounds. This measures the static
value of pre-knowing the right metrics. Result: holdout Sharpe $+0.02$,
versus B6's $-0.379$, a $+0.40$ Sharpe-unit lift.

\paragraph{B6 + PPO relay.} Skill libraries remain frozen but the
incremental PPO relay (15{,}000 steps per round) is re-enabled, for 5
rounds. This isolates the contribution of the PPO relay alone in the
absence of evolution. Result: holdout Sharpe $-1.18$, which is $-0.80$
Sharpe units worse than B6. The PPO relay alone appears actively harmful
in this configuration. The result rests on a single 5-round run; B6's
across-seed standard deviation under the train-IC rule is
$\sigma = 0.563$, so a single B6+relay observation could plausibly fall
within a couple of $\sigma$ of B6's mean. The qualitative direction (the
relay alone is not helpful) is the more reliable claim; the magnitude
$-0.80$ awaits seed replication.

\begin{table}[h]
\centering
\caption{Ablation decomposition. B6 variants run 5 outer rounds
(B6+aug additionally at 20 rounds), Agora 100 outer rounds. The
round-count confound is partially controlled by the 20-round B6+aug row;
the inner-loop budget remains uncontrolled.}
\label{tab:decomp}
\small
\begin{tabular}{lrl}
\toprule
Configuration & Holdout Sharpe & Notes \\
\midrule
B6 frozen 5rds        & $-0.379$  & Baseline ablation \\
B6 + augmented builtins & $+0.02$   & Static-metric value: $+0.40$ \\
B6 + augmented builtins, \textbf{20rds}             & $-1.96$   & More rounds, frozen libs, drift down \\
B6 + PPO relay, 5rds           & $-1.18$   & Relay alone: $-0.80$ (preliminary, $n{=}1$) \\
Agora            & $+1.872$  & Complete system \\
\midrule
$\Delta$ Total                                      & $+2.25$   & Agora $-$ B6\textsubscript{5rds} \\
$\Delta$ Static-metric        & $+0.40$   & (B6+aug\textsubscript{5rds}) $-$ B6\textsubscript{5rds} \\
$\Delta$ Agora $-$ B6+aug\textsubscript{5rds}       & $+1.85$   & Evolution + relay + round count combined \\
$\Delta$ Agora $-$ B6+aug\textsubscript{20rds}      & $+3.83$   & Same combination, controlling for round count \\
\bottomrule
\end{tabular}
\end{table}

Four findings follow from Table~\ref{tab:decomp}. (i) The static value of
knowing the two metrics is $+0.40$ Sharpe, less than 20\% of the $+2.25$
total gap. Pre-coding the two metrics into the builtin set on day one
closes only a fifth of the gap. (ii) The PPO relay alone is
harmful rather than helpful in our single observation ($-0.80$ Sharpe vs.
B6, magnitude preliminary). RL training without concurrent evolution of
the metrics that score the trained alphas does not reproduce Agora's gain.
(iii) Running B6+aug for longer makes it worse: $-1.96$ at 20 rounds vs.
$+0.02$ at 5 rounds, single seed each. With libraries frozen, the LLM
closed loop cannot update its evaluation criteria in response to new
candidates, and the alpha pool drifts toward signals that score well under
the fixed metric set without generalizing. This rules out a simple
round-count explanation for the Agora-vs-B6+aug gap. (iv) The
remaining gap (Agora minus B6+aug at 20 rounds, $+3.83$) is not
attributable to round count and is associated with metric-library
evolution beyond the two pre-loaded metrics and its interaction with
the PPO relay.

Three simple replication paths fail. Pre-coding the right metrics on day
one, adding only the PPO relay to a frozen-library system, and running
the frozen-library system longer all fall short of Agora's performance.
All B6 variants are single-seed runs and the 20-round B6+aug result
warrants seed replication; the direction of the attributions (more rounds
without library evolution does not help and may hurt; relay alone is not
helpful) is the more reliable claim than the magnitudes.

\subsection{Rolling-Window Robustness}
\label{sec:rolling}

The headline result rests on a single 5-month window (2026-01 to 2026-05).
To partially address whether it reflects a regime-specific factor loading
rather than a generalizable signal, the same Agora top-30 alphas (selected
once on train-segment IC, never re-tuned) were evaluated across six
additional 6-month windows spanning 2020 to 2025, plus the 2026 holdout.

\begin{table}[!htbp]
\centering
\caption{Rolling-window evaluation of the same Agora top-30 alphas
(selected by train-segment IC on 2014--2019; never re-trained for these
windows) across multiple regimes. \textbf{Caveat:} 2020--2025 is the test
segment that the evaluation-miner had categorical (bucketed) feedback
access to during the closed loop; the rolling-window results below are
therefore not fully out-of-sample in the strict sense. The 2026H1 row is
the only truly never-seen segment. The rolling results characterize how
the same 30 alphas perform across regimes that differ from the headline
holdout (COVID, rally, decline, multi-year structural shifts).
LS = long-short composite; Top = top-decile only; $n$ = trading days.}
\label{tab:rolling}
\footnotesize
\setlength{\tabcolsep}{4pt}
\resizebox{\linewidth}{!}{%
\begin{tabular}{lrrrrrrr}
\toprule
Window & LS Sharpe & LS Ann.\ & Top Sh. & Top Ann.\ & IC & Mono.\ & $n$ \\
\midrule
2020H1 (COVID)         & +4.585 & +1.414 & +1.637 & +0.567 & +0.1062 & +0.952 & 114 \\
2021H2 (rally)         & +0.013 & +0.002 & +1.914 & +0.316 & +0.0296 & $-$0.564 & 122 \\
2022H1 (decline)       & +1.791 & +0.413 & $-$0.853 & $-$0.258 & +0.0566 & +0.455 & 114 \\
2023H1                 & +0.541 & +0.092 & +0.410 & +0.057 & +0.0328 & $-$0.067 & 115 \\
2024H1                 & +2.900 & +0.636 & $-$0.737 & $-$0.283 & +0.0570 & +0.939 & 114 \\
2025H1                 & +2.088 & +0.414 & +2.002 & +0.431 & +0.0714 & +0.806 & 114 \\
\textbf{2026H1 (holdout)} & \textbf{+1.872} & \textbf{+0.484} & \textbf{+0.345} & \textbf{+0.065} & \textbf{+0.0865} & \textbf{+0.285} & \textbf{60} \\
\bottomrule
\end{tabular}%
}
\end{table}

The 2020--2025 windows fall inside the test segment that the
evaluation-miner observed through 4-bucket categorical feedback during the
closed loop. They are not strictly out-of-sample in the sense the 2026
holdout is. The rolling table therefore characterizes robustness across
regimes (COVID onset, rally, decline, multi-year structural shifts) more
than pure out-of-sample generalization.

Long-short Sharpe is positive in 7 of 7 windows (range $+0.013$ to
$+4.585$, mean $+1.97$, median $+1.87$). Portfolio IC is positive in 7 of 7
windows (range $+0.030$ to $+0.106$). Decile monotonicity is positive in 5
of 7 windows; it goes mildly negative in 2021 H2 (the rally regime, in
which the long-short Sharpe also collapses to $+0.013$) and slightly
negative in 2023 H1. The strongest individual window is 2020 H1 (the
COVID-onset window) at Sharpe $+4.585$, consistent with the signal
benefiting from high cross-sectional dispersion. The 2026 holdout Sharpe of
$+1.872$ sits at the median of the rolling distribution.

Two limitations qualify the rolling analysis. First, the 6-month windows do
not test seed variance; they are deterministic re-applications of the same
30 alphas. Second, the 2020--2025 windows shared categorical feedback with
the LLM closed loop and are weaker out-of-sample evidence than 2026. Only
2026 H1 is a never-seen segment in the strict sense.

\subsection{Summary}
\label{sec:results_summary}

On the 5-month never-seen holdout, Agora attains portfolio Sharpe
$+1.87$. The closest single-seed competitor is AlphaGen-PPO at
$+1.334$ (seed=42); the cross-seed B2 mean is $-0.755$ ($\sigma =
2.20$). Agora leads on portfolio IC ($+0.0894$) and decile
monotonicity ($+0.285$). Excess long-only annualized return is
$-3.29\%$, reflecting short-side concentration: the long-short
annualized return is $+48.4\%$ (G1 at $-27.39\%$, G10 at $+6.46\%$).
Cost sensitivity (Table~\ref{tab:cost_sensitivity}) holds the
long-short Sharpe between $+1.87$ and $+1.92$ across $1\times$ to
$5\times$ the default 9 bps one-way rate.

Newey-West HAC tests on daily portfolio return differences reject $H_0$
against B1 at $\alpha = 0.01$ and against B7 at $\alpha = 0.05$; the
B4 comparison is borderline ($p = 0.051$); comparisons against B2,
B3, B5, B6 do not reach conventional significance on the 91-day
window. Individual NW 95\% CIs are wide for every method
(Table~\ref{tab:nw_ci}).

The frozen-libraries ablation (B6 at $-0.379$) gives a $+2.25$ Sharpe-unit
gap to Agora. Decomposition (Section~\ref{sec:decomp}) attributes $+0.40$
to the static value of pre-loading the two LLM-discovered metrics and
$-0.80$ (preliminary, single 5-round run) to the PPO relay alone. The
remaining $+1.85$ is attributable to metric-library evolution beyond
those two metrics, the relay's interaction with evolution, and the
100-vs-5 round count. Rolling-window evaluation
(Table~\ref{tab:rolling}) reports positive long-short Sharpe in 7 of 7
windows and positive portfolio IC in 7 of 7 windows, with the 2026
holdout at the median of the distribution.

\afterpage{\FloatBarrier}

\section{Discussion}
\label{sec:discussion}

The B6 collapse to Sharpe $-0.379$ isolates the
joint contribution of F4 (persistent versioned artifact stores) and F5
(substrate-local promotion) at an upper bound of $+2.25$ Sharpe units.
Pre-loading the two LLM-discovered metrics into a frozen-library run
recovers only $+0.40$ of that gap, less than 20\%, so persistence
without an outcome-grounded promotion rule does not substitute for the
ongoing process that produces metrics. Adding the PPO relay back onto
a frozen-library substrate produces $-1.18$ Sharpe, worse than B6
alone: an asymmetric upgrade of the search procedure without a
co-evolving evaluator is harmful. The remaining $+1.85$ is associated
with further metric evolution, the relay's interaction with metric
evolution, and the 100-vs-5 round-count gap.

The four LLM-based configurations (B3, B4, B6, Agora) form a ladder
that progressively restores SJS conditions: B3 violates F1 and F4 and
produces a roulette portfolio, B4 restores no conditions and overfits
the training metric, B6 satisfies F1--F3 but disables F4--F5, Agora
satisfies F1--F5 and leads on every Table~\ref{tab:headline} dimension.
F1 (decomposition) was not directly tested by ablation. A
single-agent-with-evolution control, in which one LLM instance both
proposes alphas and evolves metrics, is the most important missing
experiment for the framework and is queued in
Appendix~\ref{app:open_issues}.

The locus of evolution was narrow. All measurable evolution occurred
in the metric library; the seven other libraries remained at builtins.
The two promoted metrics, \texttt{monotonicity\_score\_v1} and
\texttt{excess\_drawdown\_penalty\_v1}, were produced by the
agent-to-agent closed loop without a fixed target: the builtin set was
deliberately sparse, the evaluation-miner proposed candidates freely,
and the substrate-local promotion rule decided acceptance on aggregate
evidence across rounds. No single agent produced them. Both are
standard in quantitative practice.

\paragraph{Limitations.} Five limitations bound the interpretation of
these results. (i) The full 100-round Agora run was executed at one
seed at a cost of roughly 60 GPU-hours and 4{,}500 LLM calls. The B6
ablation across two seeds gives $\sigma = 0.563$, so the headline
Sharpe should be read as one realization of a stochastic process whose
variance is not fully characterized. (ii) The strongest baseline B2
swings 4.4 Sharpe units across three seeds with a cross-seed mean of
$-0.755$. The headline table reports B2 at seed=42 for parity with the
AlphaGen reporting convention; the cross-seed comparison is more
decisive than the single-seed one. (iii) The empirical promotion rule
uses training-segment Sharpe as outcome variable, which is also the
alpha-miner's optimization target; this circularity is noted
in Section~\ref{subsec:sealed-topology}. One consequence: the second
accepted metric was promoted via the agent-initiated route after its
$\texttt{avg\_pred\_corr}$ decayed below the auto-promotion threshold
by R100. (iv) The headline number rests on a single 5-month holdout;
the rolling-window analysis covers six earlier windows that shared
categorical feedback with the closed loop, so only 2026 H1 is strictly
out-of-sample. (v) The A2A decomposition is not directly ablated.
Methodological caveats, reproducibility, and a full list of queued
follow-up experiments are in Appendix~\ref{app:open_issues}.

\section{Conclusion}
\label{sec:conclusion}

Sealed Joint Search (SJS) is a framework for autonomous-discovery
problems in which the scoring function and the artifacts must
co-evolve while the external evaluator remains sealed. SJS is
specified at the information-flow level by five structural conditions
(F1--F5) and a boundary condition on the evaluator, and is
independent of any particular optimizer, language model, or operator
language.

Agora instantiates SJS as an agent-to-agent LLM system on Chinese
A-share alpha mining: five role-specialized agents, three typed
channels, eight skill libraries built on AlphaGen, and a
substrate-local promotion rule against training Sharpe. On a 91-day
2026 holdout sealed from every LLM-facing input, Agora reaches
portfolio Sharpe $+1.87$ against the strongest baseline at $+1.334$
(single seed) and $-0.755$ (cross-seed mean). Freezing the libraries
collapses Sharpe to $-0.379$ and pre-loading the two LLM-discovered
metrics into the frozen system recovers only $+0.40$ of the gap; the
static value of the metrics is therefore a small fraction of the
total. The two promoted metrics, monotonicity\_score\_v1 and
excess\_drawdown\_penalty\_v1, are standard quantitative constructs
that the substrate rediscovered through F4 and F5 from a deliberately
sparse builtin set: neither was designed in, and neither was produced
by any single agent.

Two questions remain open: whether F1 (decomposition) is necessary,
since the single-agent-with-evolution baseline is missing from the
ablation grid, and whether F1--F5 transfer to other
autonomous-discovery domains.

\afterpage{\FloatBarrier}

\appendix
\section{Implementation Details}
\label{app:implementation}

This appendix records the concrete versions, file paths,
hyperparameters, and access tables that allow Agora to be re-run.
The communication channels (A, B, C), the five-role schema, and the
four sealing enforcement points referenced below are defined in
Section~\ref{sec:method}.

\subsection{Stack and Versions}

\begin{itemize}
\item Python 3.11, PyTorch 2.8 with CUDA 12.9.
\item AlphaGen \cite{yu2023alphagen} as a vendored submodule, used both
for the operator/expression types and as B2.
\item \texttt{anthropic} SDK 0.110, \texttt{claude-sonnet-4-6} for all
nine LLM clients.
\item \texttt{stable-baselines3} 2.9 + \texttt{sb3-contrib} 2.9 for the
PPO relay and B2.
\item \texttt{gplearn} 0.4.2 for B1.
\item Single-machine: AMD Ryzen 9 + NVIDIA RTX 5090.
\end{itemize}

\subsection{Data Adapter}

OHLCV is loaded from a single RiceQuant-exported \texttt{.ftr} file
(2{,}782 instruments $\times$ 3{,}493 trading days, post-adjusted).
Universe membership is loaded from a daily-binary parquet
(\texttt{zz1000\_components.parquet}). VWAP is reconstructed as
$(H{+}L{+}C)/3$ because RiceQuant turnover is not adjusted. Suspension days
($\mathrm{volume}=0$) have all OHLC values overwritten with NaN.
Limit-up days are masked out of portfolio entries (a stock cannot
realistically be bought at the print).

The four-segment split used throughout: train 2014-10 to 2019-12
(1{,}277 days), test 2020-01 to 2025-12 (1{,}461 days), holdout
2026-01 to 2026-05 (91 days), with an internal validation slice carved
from the train tail for the PPO relay.

\subsection{The PPO Relay}

When enabled (it is enabled in the Agora run), each round runs an
incremental 15{,}000-step MaskablePPO training continuation seeded from
the previous round's policy weights. The action space and observation
space are identical to AlphaGen's published configuration. The relay is
disabled in B6 and B7 because it would otherwise mutate the
operator/network/RL-algorithm libraries indirectly through training.

A separate \texttt{force\_segment="train"} flag is passed through to the
alpha-miner's evaluation calls, so that no test-segment numerical value
can enter the PPO reward path. The PPO test-IC computation writes to a
separate audit file (\texttt{top\_alphas\_audit.json}) that is read only
by the offline analysis path; the live \texttt{top\_alphas.json}
consumed by the LLM stack contains training IC only.

\subsection{Residual Leakage Audit}
\label{app:leakage}

The four-segment split alone is necessary but not sufficient. Five
distinct paths along which test or holdout values could leak into the
LLM-facing closed loop were audited and sealed:

\begin{enumerate}
\item \textbf{\texttt{test\_ic} / \texttt{test\_rank\_ic} stripped from
every LLM-facing path.} \texttt{AlphaPoolAdapter} (which seeds the alpha
miner from the PPO pool) and \texttt{alpha\_miner.propose} both filter
these fields out of the candidate metadata.
\item \textbf{Cross-round metric feedback uses categorical buckets only.}
The \texttt{\_build\_metrics\_feedback} routine reports ``far above /
above / near / far below'' rather than numerical Sharpe, IC, annualized
return, or drawdown.
\item \textbf{Test-segment backtest feedback in read-only categorical form.}
The evaluation-miner calls \texttt{simple\_backtest\_layered} on the
test segment (2020--2025) to check whether a proposed alpha holds up
beyond the training window. The numerical Sharpe value
is never passed to the LLM context: \texttt{\_build\_metrics\_feedback}
converts it to one of four categorical labels (``far above'' $[>+1.5]$,
``above'' $[+0.5, +1.5]$, ``near'' $[-0.5, +0.5]$, ``far below''
$[<-0.5]$ relative to a fixed Sharpe threshold of $+0.5$) before
inclusion in the agent's user message.

\textbf{Residual leakage risk.}
No numerical value from this segment enters the LLM
context as a learning signal, but this does not imply the test segment is clean. Over 100 rounds
$\times$ ${\sim}45$ LLM calls, the evaluation-miner accumulates a
history of which metric proposals correlate with ``above'' or ``far
above'' labels on the 2020--2025 window. This is a soft optimization
signal. An information-theoretic upper bound: each round delivers at
most $\log_2(4) = 2$ bits of test-segment information per alpha
evaluated; across 100 rounds with ${\sim}10$ alpha evaluations per
round, the accumulated signal is at most ${\sim}2{,}000$ bits. The
effective signal is much weaker because (a) the four-bucket
discretization is coarse (each bucket spans a Sharpe range of
$\geq 1.0$), (b) the metric-promotion decision is governed by
\texttt{avg\_pred\_corr} against \emph{train}-segment Sharpe, not
test-segment feedback, so the test-segment signal cannot directly
pass the promotion threshold, and (c) the evaluation-miner's message
history is reset at the start of each round, limiting within-context
accumulation.

The practical consequence is that the holdout segment (2026-01 to
2026-05) is the only segment fully clean for the composite system,
including the metric-selection sub-process. The test segment
(2020--2025) is a validation set the metric-evolution process has soft
access to via categorical feedback. This appendix does not claim the
test segment is out-of-sample for metric selection; only that the
holdout segment is. A stronger guard, blocking test-segment backtest
calls entirely and relying only on train-segment feedback for the
evaluation-miner, would remove this residual concern.
\item \textbf{Metric \texttt{avg\_pred\_corr} uses train-segment Sharpe.}
The auto-management correlation is computed as the Pearson correlation
between the metric's score and the \emph{train}-segment Sharpe of the
alpha being scored, accumulated across all rounds in which the metric
was used. Test-segment Sharpe is never included in this computation.
The promotion threshold is $|\texttt{avg\_pred\_corr}| > 0.3$ with
demotion at $0.05$ and a minimum of $n_{\mathrm{obs}}=3$ before either
direction can fire.
\item \textbf{The PPO test-IC computation writes to a separate audit
file.} The \texttt{top\_alphas.json} consumed by the LLM stack contains
only training IC; \texttt{top\_alphas\_audit.json} contains test IC and
is read only by the offline analysis path.
\end{enumerate}

These five paths correspond to the four sealing enforcement points
described in Section~\ref{sec:method}: the Wiki access-role filter
(point 1), Channel B input zeroing of numerical test fields
(points 2, 3), the \texttt{force\_segment="train"} alpha-miner flag
(point 4), and the four-bucket categorical conversion for test feedback
(point 3). The Wiki access-role table
(Appendix~\ref{app:wiki_roles}) governs which fields each role can
read; the holdout segment is loaded by the \texttt{ExprEvaluator}
singleton only when the demo exits, never during the closed loop.

\subsection{Wiki Access-Role Matrix}
\label{app:wiki_roles}

Channel C is the LLM Wiki. Read access is filtered per role at the
adapter layer; the matrix below records which fields each role can read.

\begin{center}
\small
\begin{tabular}{lccccc}
\toprule
Field / channel & RR & A.Miner & A.Eval & E.Miner & FME \\
\midrule
\texttt{train\_ic}                                     & \checkmark & \checkmark & \checkmark & \checkmark & \checkmark \\
\texttt{valid\_sharpe}                                 & --         & --         & --         & \checkmark & \checkmark \\
\texttt{test\_*} wiki fields (numerical)               & --         & --         & --         & --         & --         \\
Test-segment backtest (categorical bucket only)$^{*}$  & --         & --         & --         & \checkmark & \checkmark \\
\texttt{holdout\_*} (all holdout fields)               & --         & --         & --         & --         & --         \\
\bottomrule
\end{tabular}
\end{center}

\noindent$^{*}$The evaluation-miner and factor-metrics-evaluator call
\texttt{simple\_backtest\_layered} on the test segment (2020--2025) but
receive only a four-bucket categorical label (``far above / above / near
/ far below'' a Sharpe threshold of $+0.5$), not the numerical Sharpe
value. The numerical test-segment fields (\texttt{test\_ic},
\texttt{test\_sharpe}, etc.) are never written to the wiki and never
appear in any LLM-facing message. See Appendix~\ref{app:leakage},
point~3 for the residual in-context learning caveat.

RR = research\_report; A.Miner = alpha\_miner; A.Eval = alpha\_evaluator;
E.Miner = evaluation\_miner; FME = factor\_metrics\_evaluator.

\subsection{Wiki Write-Topology}
\label{app:wiki_write}

The table below extends the access-role matrix with write permissions,
showing which agents write to which wiki sections and which agents
subsequently read those sections. This is the information-flow graph
of the LLM Wiki.

\begin{center}
\small
\begin{tabular}{lllp{4.2cm}}
\toprule
Wiki section & Writer & Readers & Cross-agent flow \\
\midrule
\texttt{factors/}     & orch (post-backtest) & RR, A.Miner, A.Eval, E.Miner, FME & Per-alpha results published once a candidate passes the train-segment evaluator panel; downstream agents see expression, $\texttt{train\_ic}$, evaluator score, and a categorical (not numerical) test-segment label. \\
\texttt{failures/}    & orch (on reject) & RR, A.Miner, A.Eval & Records why an alpha was rejected and which rubric fired; lets the next round's alpha-miner avoid the same failure mode. \\
\texttt{verdicts/}    & orch (post-aggregation) & A.Miner, A.Eval, FME & Three-instance evaluator panel reports and rationales (per-evaluator), aggregated by the orchestrator into briefs without voting; makes the panel's disagreement visible to all readers. \\
\texttt{evaluators/}  & orch (per round) & RR, E.Miner, FME & Snapshots of the current \texttt{metric\_set} and per-metric usage; lets the meta-evaluator panel see what scoring formula was in use. \\
\texttt{reports/}     & orch (RR draft) & A.Miner, E.Miner & The research-report agent's per-round advice, summarised into wiki form. \\
\texttt{portfolios/}  & orch (post-composite) & A.Miner, E.Miner, FME & Composite-portfolio metrics (z-score equal-weight); allows the next round's miners to see how their alphas combine. \\
\texttt{concepts/}    & orch (lint pass) & RR, A.Miner & Operator-frequency clusters (e.g.\ ``$\mathrm{Rank}$ used $489\times$, $\mathrm{Corr}$ $349\times$''); used by RR to suggest under-explored concept directions. \\
\texttt{regimes/}     & orch (lint pass) & RR, E.Miner & Cross-section dispersion / volatility regime labels per round. \\
\texttt{sources/}     & RR (ingest) & RR & Free-form research notes; only RR writes and reads (not used by miners). \\
\bottomrule
\end{tabular}
\end{center}

\noindent The orchestrator is the sole writer for all sections except
\texttt{sources/}; agents do not write directly to the wiki.
The categorical test-segment label and
access-role filtering described above (Appendix~\ref{app:leakage},
point~3) are enforced at write time, not at read time, so a future
agent that accidentally requested a numerical test field would not
receive it.

\subsection{Agent Specialization Evidence}

The three-instance \texttt{alpha\_evaluator} panel and three-instance
\texttt{factor\_metrics\_evaluator} panel are not stylistic duplicates:
each instance has its own LLM client, its own message history, and reads
disjoint slices of the wiki via the access-role table. Evaluator panels
do not vote; each instance writes a report, and the orchestrator
aggregates the reports into the cross-round briefs that flow on
Channel B.

Across R1--R100 of the Agora run, the
three alpha-evaluator instances produced \texttt{specific\_findings}
lists whose lengths differed by 3 or more in 1\% of rounds. That figure excludes cases where lists are
similar in length but differ in content, and it does not capture
2--1 verdict splits at the per-alpha level. The research\_report and
alpha\_miner agents draw from disjoint skill libraries
(\texttt{topic\_library} vs.\ four miner libraries), and their
\texttt{system\_prompt}s are non-overlapping.

\subsection{Reproducibility}
\label{app:reproducibility}

The release accompanying this paper will include: (i) the full A2A
codebase, including the orchestrator, the eight skill libraries with
their builtin sets, and the LLM Wiki schema; (ii) all seven baseline
implementations, sharing a common evaluation harness with Agora;
(iii) the data adapter and the four-segment time split; (iv) all
training logs, including the \texttt{registry.json} and
\texttt{stats.json} snapshots that record the per-round state of
every library. The trained LLM context itself is not released, as it
is provider-specific.

\subsection{Open Issues}
\label{app:open_issues}

The following experiments are deferred to future work or a
camera-ready revision.

\begin{enumerate}
\item \textbf{Full 100-round multi-seed Agora.} The full system has
been run at one seed only. A second full seed (approximately 60
GPU-hours) would bound the full-system seed variance.
\item \textbf{100-round B6 and B6+augmented-builtins runs.} The B6
ablation variants ran for 5 outer rounds; Agora ran for 100. A
100-round B6+aug run (roughly 30 hours) would close the round-count
confound on the $+0.40$ ``static value of two metrics'' attribution
and the $-0.80$ ``PPO relay alone'' attribution. Both attributions
are expected to hold in direction; the magnitudes are not yet
characterized at the matching round budget.
\item \textbf{Multi-seed replication of the decomposition ablations.}
The B6+aug 20-round and B6+relay 5-round results each rest on a
single seed. Replication at additional seeds (around 1.5--2 hours
each) would tighten the magnitude of the $-0.80$ ``PPO relay alone
is harmful'' attribution, which is currently preliminary.
\item \textbf{Single-agent-with-evolution baseline.} A direct
ablation of F1 (decomposition) is the most important missing
empirical test of SJS. One LLM instance that both proposes alphas
and evolves metrics, without the A2A decomposition, would isolate
the contribution of F1 from F4--F5. We have not run it.
\item \textbf{Operator, network, and reward library evolution.} Seven
of the eight libraries did not evolve in the reported run. Whether
operator, network, or reward library evolution adds further lift
under different prompts, exploration policies, or longer horizons is
open. The infrastructure supports it; the budget for this draft did
not.
\item \textbf{Replication on non-Chinese universes.} The data adapter
is dataset-agnostic; porting to S\&P 500 or NASDAQ requires only the
adapter. We have not performed this experiment.
\item \textbf{Generalization of SJS beyond alpha mining.} The
information-flow contract is plausibly domain-agnostic; the
cost-effective realization (which LLM, which substrate, which
promotion outcome variable) is domain-specific. Testing F1--F5 on
autonomous theorem proving, autonomous experimental design, or
autonomous code synthesis is the direct check.
\item \textbf{Factor decomposition and capacity analysis.}
Deployment-relevant questions (factor-adjusted alpha relative to
\cite{fama1993,fama2015,carhart1997}, ADV-based capacity,
short-borrow availability, multiple-testing correction following
\cite{harvey2016,white2000}) require new data infrastructure
outside the present evaluation harness.
\end{enumerate}

\afterpage{\FloatBarrier}

\section{Baseline Implementation Details}
\label{app:baseline_details}

Each baseline shares a single evaluation harness: a $(T \times N)$
factor panel plus a segment specification go in, and a dictionary of
metrics is returned, computed by the same
\texttt{simple\_backtest\_layered} engine that scores Agora's alphas.
Hyperparameters below are sufficient for reproduction.

\subsection{B1 (GP / gplearn)}

The official AlphaGen GP reference implementation is adapted unchanged
with respect to the algorithm, replacing only the data adapter
(CSI 1000 in place of CSI 300, 5d open-to-open target in place of the
original 20d close-to-close) and the evaluation route.
\begin{itemize}[leftmargin=*]
\item Population size: 500.
\item Generations: 20.
\item Init depth: $(2, 6)$.
\item Tournament size: 100.
\item Crossover / sub-tree mutation / hoist mutation / point mutation
$= 0.3 / 0.1 / 0.01 / 0.1$.
\item Stopping criterion: 1.0 (i.e., never; all 20 generations are run).
\item Token-length cap: 20 (longer trees are assigned fitness $-1.0$).
\item Fitness: per-day rank IC averaged over the train segment, computed
by \texttt{calc.calc\_single\_IC\_ret} (the same call AlphaGen uses).
\item Seed: 42.
\item Top-K extraction: \texttt{Counter(cache)}.\allowbreak\texttt{most\_common(top\_k)}.
\end{itemize}

\paragraph{Parser bypass.} GP search builds expressions by string-tree
concatenation; some resulting expressions are not accepted by
AlphaGen's \texttt{parse\_expression} routine (for example,
\texttt{EMA(Constant(-2.0), 20)} is rejected because its operand is not
a ``featured'' expression). For evaluation, the parser is bypassed and
the expression object is reconstructed directly via
\texttt{eval(key)}, the same path the GP fitness function uses. This
affects only how the search-time expression is recovered for
evaluation; it does not change what GP searched over.

\subsection{B2 (AlphaGen pure PPO)}

Agora's published training script
(\texttt{scripts/train\_alphagen\_ppo.py}) is invoked with the
LLM stack disengaged.

\begin{itemize}[leftmargin=*]
\item Total timesteps: 500{,}000.
\item Pool capacity: 30 (matches the top-K used elsewhere).
\item Policy: LSTM shared net, 2 layers, $d_{\mathrm{model}}=128$,
dropout 0.1.
\item Algorithm: MaskablePPO.
\item Reward: training-segment IC against
$\mathrm{Ref}(\mathrm{open}, -6) / \mathrm{Ref}(\mathrm{open}, -1) - 1$.
\item Device: \texttt{cuda:0}.
\item Seed: 42.
\item Evaluation: load \texttt{top\_alphas.json}; route through the
common evaluation harness, which uses the same \texttt{evaluate\_exprs}
parser path Agora uses.
\end{itemize}

Training wall-clock: roughly 5 hours on RTX 5090.

\subsection{B3 (Single LLM, one-shot)}

\begin{itemize}[leftmargin=*]
\item Model: \texttt{claude-sonnet-4-6}.
\item Temperature: 0.9 (matches Agora's \texttt{alpha\_miner}).
\item Max tokens: 8000.
\item System prompt: a description of the alphagen operator set, the 5d
open-to-open target, and the requirement that alphas not reference
future values. Identical operator vocabulary to Agora.
\item User prompt: ``design $N=50$ alpha factor expressions, output a
JSON object with key \texttt{alphas}.''
\item No iteration, no feedback.
\end{itemize}

Survivors are deduped by normalized expression string and all unique
ones are fed through the common evaluation harness (truncated to top-30
if needed).

\subsection{B4 (Single LLM, iterative)}

Same model and prompt scaffold as B3, but iterated:

\begin{itemize}[leftmargin=*]
\item 10 rounds.
\item 20 alphas requested per round.
\item Feedback per round: top-3 best and bottom-3 worst alphas (by
training IC) from prior rounds, with their rationales, fed back into
the user message.
\item Train IC fitness computed identically to B1's fitness function
(via \texttt{batch\_pearsonr} on \texttt{evaluate\_alpha} output, the
same call AlphaGen's PPO uses for its reward). A separate train
calculator is built once and reused across rounds.
\item Final selection: top-30 from the union of all 10 rounds, ranked by
train IC, fed through the common evaluation harness.
\end{itemize}

\subsection{B5 (Alpha101)}

The OHLCV-only subset of WorldQuant's 101 formulas is implemented:
$\alpha_1$ through $\alpha_{30}$. Each is computed as a single-pass
panel operation, with rolling correlations and cross-sectional rank
operations applied per day.

Custom variants (sector-neutralized or industry-bucketed
alphas, both of which require features unavailable for the CSI 1000
universe) are not substituted. Among $\alpha_{15}$ through
$\alpha_{30}$, those that involve \texttt{indneutralize} or
\texttt{IndClass} are skipped; the surviving subset is what is reported.

\subsection{B6 (Frozen libraries ablation)}

The B6 isolated environment is constructed by:

\begin{enumerate}
\item Cloning Agora's repository to a sibling directory with \texttt{wiki/}
emptied, \texttt{runs/} emptied, and every skill library's
\texttt{discovered/} contents removed and \texttt{registry.json} reset
so only the builtin sets remain (64 builtin skills total).
\item Monkey-patching every library class's \texttt{add},
\texttt{modify}, \texttt{promote}, \texttt{demote}, and
\texttt{auto\_manage} methods to no-op (returning a failure tuple of
the appropriate arity) before importing the orchestrator.
\item Setting \texttt{enable\_incremental\_ppo = False} (the PPO relay
would otherwise mutate operator/network/RL-algorithm libraries
indirectly through training, breaking the ablation's framing as
\emph{frozen} libraries; this also introduces an additional confound in
the Agora-vs-B6 attribution that is discussed in
Appendix~\ref{app:open_issues}).
\item Running the orchestrator for \textbf{5 outer rounds}.
Wall-clock: approximately 1.5 hours per run.
\item Reading the resulting \texttt{holdout\_report.json} and routing
the top-30 alphas (selected by \textbf{train-segment IC}, matching the
selection rule used for every other method) through the common
evaluation harness.
\end{enumerate}

The configuration is otherwise identical to Agora.

\paragraph{Round-count confound.}
B6 was run for 5 outer rounds; Agora was run for 100. The round count is a real confound: even with frozen libraries, additional rounds accumulate more alpha candidates and evaluator feedback, which could lift B6's Sharpe. A 20-round or
100-round B6 is queued in Appendix~\ref{app:open_issues}.
\subsection{B7 (Random search)}

\begin{itemize}[leftmargin=*]
\item 3{,}000 random expression trees, max depth 4.
\item Sampled uniformly from the alphagen operator set with the same
constants and delta-times AlphaGen uses
($\{1, 5, 10, 20, 40\}$ for $\Delta t$).
\item Fitness: same as B4 (per-day Pearson IC on
\texttt{evaluate\_alpha} output).
\item Top-30 by signed IC; fed through the common evaluation harness.
\end{itemize}

\subsection{Significance Testing}

The significance tests run on each baseline's per-alpha holdout Sharpe
distribution and on the daily holdout NAV series:
\begin{enumerate}
\item \textbf{(Primary) Newey-West HAC $t$-test} on the daily difference
in composite portfolio log-returns
$d_t = r^{\mathrm{Agora}}_t - r^{X}_t$. The Newey-West sandwich
variance estimator is computed directly:
$\sigma^2_{\mathrm{NW}} = \gamma_0 + 2 \sum_{L=1}^{5} (1 - L/(L_{\max}+1)) \gamma_L$
with Bartlett kernel and $L_{\max}=5$, where $\gamma_L$ is the lag-$L$
autocovariance of $d_t$ \cite{newey1987}. The annualized Sharpe SE
uses the delta-method approximation
$\mathrm{SE}(\mathrm{Sharpe}_{\mathrm{ann}})
\approx \mathrm{SE}(\mu) / \sigma \sqrt{252}$, ignoring the variance
of $\sigma$. A manual implementation is used rather than a library
routine so the estimator is fully auditable. The reported
$t$-statistic, one-sided $p$-value ($H_0: \mathbb{E}[d_t] \leq 0$),
and Newey-West 95\% CI on the annualized Sharpe in
Table~\ref{tab:nw_ci} all come from this implementation.
\item \textbf{(Secondary, descriptive) Two-sample bootstrap}
(10{,}000 resamples, NumPy default RNG seeded with 42) on the difference
of medians of the two per-alpha Sharpe distributions; reports the
observed difference, 2.5\%/97.5\% percentiles as the 95\% CI, and a
two-sided $p$-value. Reported for distributional characterization only;
not a valid portfolio-level test due to cross-sectional correlation.
\item \textbf{(Secondary, descriptive) One-sided Mann--Whitney $U$ test}
of Agora versus the baseline.
\end{enumerate}

These two secondary tests are reported in Table~\ref{tab:sigtests}
alongside the primary NW HAC results.

\afterpage{\FloatBarrier}

\section{Source Code: Two Promoted Metrics}
\label{app:metric_source}

This appendix lists the full source of the two metrics promoted
from \texttt{trial} to \texttt{accepted} during the 100-round Agora
run. Both files are written by the \texttt{evaluation\_miner} agent at
runtime (R21 and R50, respectively), then validated through the AST
sandbox plus dry-run before entering the live closed loop.

Two facts are verifiable from the listings: (i)~the metrics implement
standard quant practice (decile-monotonicity Spearman correlation;
excess-drawdown nonlinear penalty plus turnover-cost discount); they
are not novel discoveries. (ii)~Each fits in fewer than 100 lines and
uses only \texttt{numpy} and \texttt{scipy.stats}, which the AST
sandbox whitelist permits.

\subsection{\texttt{monotonicity\_score\_v1.py} (proposed at R21)}

Promotion route: agent-initiated, meta-evaluator-approved. The metric's
\texttt{avg\_pred\_corr} on train-segment Sharpe was $+0.47$ at
promotion in R21, above the auto-management threshold of $|0.3|$ at
that moment but with $n_{\mathrm{obs}}$ still small. The cumulative
\texttt{avg\_pred\_corr} drifted to $+0.158$ by R100 (below the $0.3$
threshold but above the $0.05$ demotion floor), reflecting that the
metric became less predictive in later rounds as the alpha pool
composition shifted. Because demotion requires
$|\texttt{avg\_pred\_corr}| < 0.05$ with $n_{\mathrm{obs}} \geq 3$,
the metric remained \texttt{accepted} through R100. Promotion was
agent-initiated; the \texttt{avg\_pred\_corr} at promotion was
supportive but not the primary evidence.

\begin{verbatim}
"""
name: monotonicity_score_v1
status: accepted
version: 1
description: Spearman monotonicity score across 5 quantile groups,
             penalizing factors whose monotonicity is unstable over
             time (high std across days).
rationale: standard IC measures full cross-section correlation but
           cannot distinguish "correlated but non-monotonic
           bucketing" (a beta-loading factor) from "strict
           monotonic bucketing with modest IC" (a true alpha).
           Particularly useful in low-dispersion regimes where
           short-side blowups can produce spurious long-short
           spreads.
"""

import numpy as np
from scipy import stats

def compute(factor_values, future_returns, **kwargs):
    factor_values = np.asarray(factor_values, dtype=float)
    future_returns = np.asarray(future_returns, dtype=float)

    if factor_values.ndim == 1:
        factor_values = factor_values.reshape(1, -1)
        future_returns = future_returns.reshape(1, -1)

    T, N = factor_values.shape
    if N < 20:
        return 0.0

    n_groups = 5
    mono_scores = []

    for t in range(T):
        fv = factor_values[t]
        fr = future_returns[t]

        valid = np.isfinite(fv) & np.isfinite(fr)
        if valid.sum() < n_groups * 4:
            continue

        fv_v = fv[valid]
        fr_v = fr[valid]

        sorted_idx = np.argsort(fv_v)
        group_size = len(sorted_idx) // n_groups

        group_returns = []
        for g in range(n_groups):
            if g < n_groups - 1:
                idx = sorted_idx[g * group_size:(g + 1) * group_size]
            else:
                idx = sorted_idx[g * group_size:]
            group_returns.append(np.mean(fr_v[idx]))

        group_returns = np.array(group_returns)
        group_ranks = np.arange(1, n_groups + 1)

        corr, _ = stats.spearmanr(group_ranks, group_returns)
        if np.isfinite(corr):
            mono_scores.append(corr)

    if len(mono_scores) == 0:
        return 0.0

    avg_mono = float(np.mean(mono_scores))

    # Stability penalty: factors with high day-over-day variance in
    # monotonicity get downweighted.
    if len(mono_scores) > 2:
        std_mono = float(np.std(mono_scores))
        stability_penalty = max(0.0, std_mono - 0.3) * 0.5
        avg_mono = avg_mono - stability_penalty

    return float(np.clip(avg_mono, -1.0, 1.0))
\end{verbatim}

\subsection{\texttt{excess\_drawdown\_penalty\_v1.py} (proposed at R50)}

Promotion route: empirical (auto-managed). The metric's
\texttt{avg\_pred\_corr} on train-segment Sharpe was $+0.557$ at
promotion in R50 ($n_{\mathrm{obs}}=26$), well above the $|0.3|$
threshold; the cumulative \texttt{avg\_pred\_corr} at R100 remained
above threshold throughout. This is the cleanest case of empirical
validation in the metric library.

\begin{verbatim}
"""
name: excess_drawdown_penalty_v1
status: accepted
version: 1
description: Top-quantile excess drawdown vs. cross-sectional mean,
             with nonlinear penalty when drawdown < -20% and a
             turnover-cost discount.
rationale: IC and IR cannot distinguish "high IC masked by high
           turnover producing negative net excess" from "high IC
           with low turnover and positive net excess". Excess
           drawdown is a core dimension for separating real alpha
           from market-beta exposure, particularly in
           low-dispersion regimes where transaction costs are a
           large fraction of gross excess.
"""

import numpy as np
from scipy.stats import spearmanr

def compute(factor_values, future_returns, **kwargs):
    fv = np.array(factor_values)
    fr = np.array(future_returns)
    T, N = fv.shape
    if T < 10 or N < 10:
        return 0.0

    n_quantiles = 5
    top_returns = []
    bottom_returns = []
    market_returns = []
    turnovers = []
    prev_top_mask = None

    for t in range(T):
        f = fv[t]
        r = fr[t]
        valid = np.isfinite(f) & np.isfinite(r)
        if valid.sum() < n_quantiles * 2:
            continue
        f_v = f[valid]
        r_v = r[valid]
        n_v = len(f_v)
        k = max(1, n_v // n_quantiles)
        sorted_idx = np.argsort(f_v)
        top_idx = sorted_idx[-k:]
        bottom_idx = sorted_idx[:k]
        top_ret = np.mean(r_v[top_idx])
        bottom_ret = np.mean(r_v[bottom_idx])
        mkt_ret = np.mean(r_v)
        top_returns.append(top_ret)
        bottom_returns.append(bottom_ret)
        market_returns.append(mkt_ret)

        # Turnover proxy: change in the set of names in the top quantile.
        valid_indices = np.where(valid)[0]
        top_global = set(valid_indices[top_idx])
        if prev_top_mask is not None:
            overlap = len(top_global & prev_top_mask)
            turnover = 1.0 - overlap / max(len(top_global), 1)
        else:
            turnover = 1.0
        turnovers.append(turnover)
        prev_top_mask = top_global

    if len(top_returns) < 5:
        return 0.0

    top_arr = np.array(top_returns)
    mkt_arr = np.array(market_returns)
    turnover_arr = np.array(turnovers)

    excess = top_arr - mkt_arr

    # One-side cost charge of 10 bps per unit turnover.
    cost_per_day = turnover_arr * 0.001
    net_excess = excess - cost_per_day

    cum_excess = np.cumsum(net_excess)
    running_max = np.maximum.accumulate(cum_excess)
    drawdown = cum_excess - running_max
    max_drawdown = np.min(drawdown)

    ann_factor = 250.0 / len(net_excess)
    ann_excess = np.sum(net_excess) * ann_factor

    # Piecewise drawdown penalty.
    if max_drawdown < -0.30:
        drawdown_penalty = -0.5
    elif max_drawdown < -0.20:
        drawdown_penalty = -0.25 + (max_drawdown + 0.20) * 1.25
    elif max_drawdown < -0.10:
        drawdown_penalty = -0.05 + (max_drawdown + 0.10) * 0.5
    else:
        drawdown_penalty = 0.0

    avg_turnover = np.mean(turnover_arr[1:])
    turnover_penalty = -max(0.0, avg_turnover - 0.3) * 0.3

    # Squashed excess-return component.
    excess_score = float(np.tanh(ann_excess * 5.0)) * 0.5

    score = excess_score + drawdown_penalty + turnover_penalty
    return float(np.clip(score, -1.0, 1.0))
\end{verbatim}

\paragraph{Assessment.} Both implementations are standard
quant practice that any experienced researcher could write in 30--60
minutes. What the listing shows is the closed-loop process: the \texttt{evaluation\_miner} (i)~identified gaps in
the existing metric set, (ii)~drafted the implementations as Python
source, (iii)~had them validated by the AST sandbox plus dry-run, and
(iv)~accumulated empirical predictive correlation across rounds. For
\texttt{excess\_drawdown\_penalty\_v1}, the empirical accumulation
cleared the $0.3$ threshold at promotion and stayed above it through
R100. For \texttt{monotonicity\_score\_v1}, the
\texttt{avg\_pred\_corr} was above the threshold at promotion
($+0.47$ at R21) but slid to $+0.158$ by R100; promotion was
agent-initiated and meta-evaluator-approved rather than driven by
empirical accumulation, and the post-promotion drift means the
``discovery'' framing should be weighted accordingly.

\afterpage{\FloatBarrier}

\end{document}